\documentclass[conference]{IEEEtran}
\IEEEoverridecommandlockouts
\usepackage{svg}

\usepackage{cite}
\usepackage{amsmath,amssymb,amsfonts}
\usepackage{algorithm}
\usepackage{algorithmic}
\usepackage{graphicx}
\usepackage{textcomp}
\usepackage{lettrine}
\usepackage{booktabs, threeparttable}
\usepackage{float} 
\usepackage{xcolor}
\usepackage{hyperref}
\usepackage{fancyhdr}

\usepackage[font=small,labelfont=bf]{caption}
\usepackage{makecell}
\newcolumntype{C}{>{\centering\arraybackslash}p{2cm}}
\usepackage{soul}

\usepackage[normalem]{ulem}
\usepackage[font=small,labelfont=bf]{caption}

\hypersetup{
    colorlinks=true,       
    linkcolor=blue,        
    citecolor=blue,        
    urlcolor=black          
}

\def\BibTeX{{\rm B\kern-.05em{\sc i\kern-.025em b}\kern-.08em
    T\kern-.1667em\lower.7ex\hbox{E}\kern-.125emX}}
\begin{document}
\title{MMGT: Motion Mask Guided Two-Stage Network for Co-Speech Gesture Video Generation\\
{\footnotesize}
}

\author{
Siyuan~Wang$^*$,
Jiawei~Liu$^*$,~\IEEEmembership{Member,~IEEE,}
Wei~Wang$^\dagger$,
Yeying Jin,~\IEEEmembership{Member,~IEEE,}\\
Jinsong Du,
Zhi Han,~\IEEEmembership{Member,~IEEE}
\thanks{This work was supported in part by the National Natural Science Foundation of China under Grant U23A20343; Liaoning Provincial ``Selecting the Best Candidates by Opening Competition Mechanism" Science and Technology Program under Grant 2023JH1/10400045}
\thanks{Siyuan Wang are with Shenyang Institute of Automation, Chinese Academy of Sciences, Shenyang 110016, P. R. China, Liaoning Liaohe Laboratory, the Key Laboratory on Intelligent Detection and Equipment Technology, Shenyang 110169, P. R. China, and also with the University of Chinese Academy of Sciences, Beijing 100049, P. R. China (e-mail: wangsiyuan@sia.cn).}
\thanks{Jiawei Liu, Wei Wang, and Jinsong Du are with Shenyang Institute of Automation, Chinese Academy of Sciences, Shenyang 110016, P. R. China, Liaoning Liaohe Laboratory, the Key Laboratory on Intelligent Detection and Equipment Technology, Shenyang 110169, P. R. China (e-mail: liujiawei@sia.cn; wangwei2@sia.cn; jsdu@sia.cn).}
\thanks{Yeying Jin was with National University of Singapore (NUS); (e-mail: jinyeying@u.nus.edu).}
\thanks{Zhi Han was with the State Key Laboratory of Robotics, Shenyang Institute of Automation, Chinese Academy of Sciences, Shenyang 110016, P. R. China (e-mail: hanzhi@sia.cn).}
\thanks{$^*$These authors contributed equally to this work.}
\thanks{$^\dagger$Corresponding author: Wei Wang (email: wangwei2@sia.cn).}
}
\maketitle

\IEEEpubid{%
\begin{minipage}{\textwidth}\ \\[35pt] 
\centering\footnotesize
Copyright © 2025 IEEE. Personal use of this material is permitted. \\
However, permission to use this material for any other purposes must be obtained 
from the IEEE by sending an email to pubs-permissions@ieee.org.
\end{minipage}}

\vspace{-1em} 
\begin{abstract}
\textbf{Co-Speech Gesture Video Generation aims to generate vivid speech videos from audio-driven still images, which is challenging due to the diversity of body parts in terms of motion amplitude, audio relevance, and detailed features. Relying solely on audio as the control signal often fails to capture large gesture movements in videos, resulting in more noticeable artifacts and distortions. Existing approaches typically address this issue by adding extra prior inputs, but this can limit the practical application of the task. Specifically, we propose a Motion Mask-Guided Two-Stage Network (MMGT) that uses audio, along with motion masks and pose videos generated from the audio signal, to jointly generate synchronized speech gesture videos. In the first stage, the Spatial Mask-Guided Audio2Pose Generation (SMGA) Network generates high-quality pose videos and motion masks from audio, effectively capturing large movements in key regions such as the face and gestures. In the second stage, we integrate Motion Masked Hierarchical Audio Attention (MM-HAA) into the Stabilized Diffusion Video Generation model, addressing limitations in fine-grained motion generation and region-specific detail control found in traditional methods. This ensures high-quality, detailed upper-body videos with accurate textures and motion. 
Evaluations demonstrate improvements in video quality, lip-sync, and hand gestures. The model and code are available at (\href{https://github.com/SIA-IDE/MMGT}{\color{blue}{https://github.com/SIA-IDE/MMGT}}).}

\end{abstract}


\begin{figure}[t]
\centerline{%
  \includegraphics[width=0.48\textwidth]{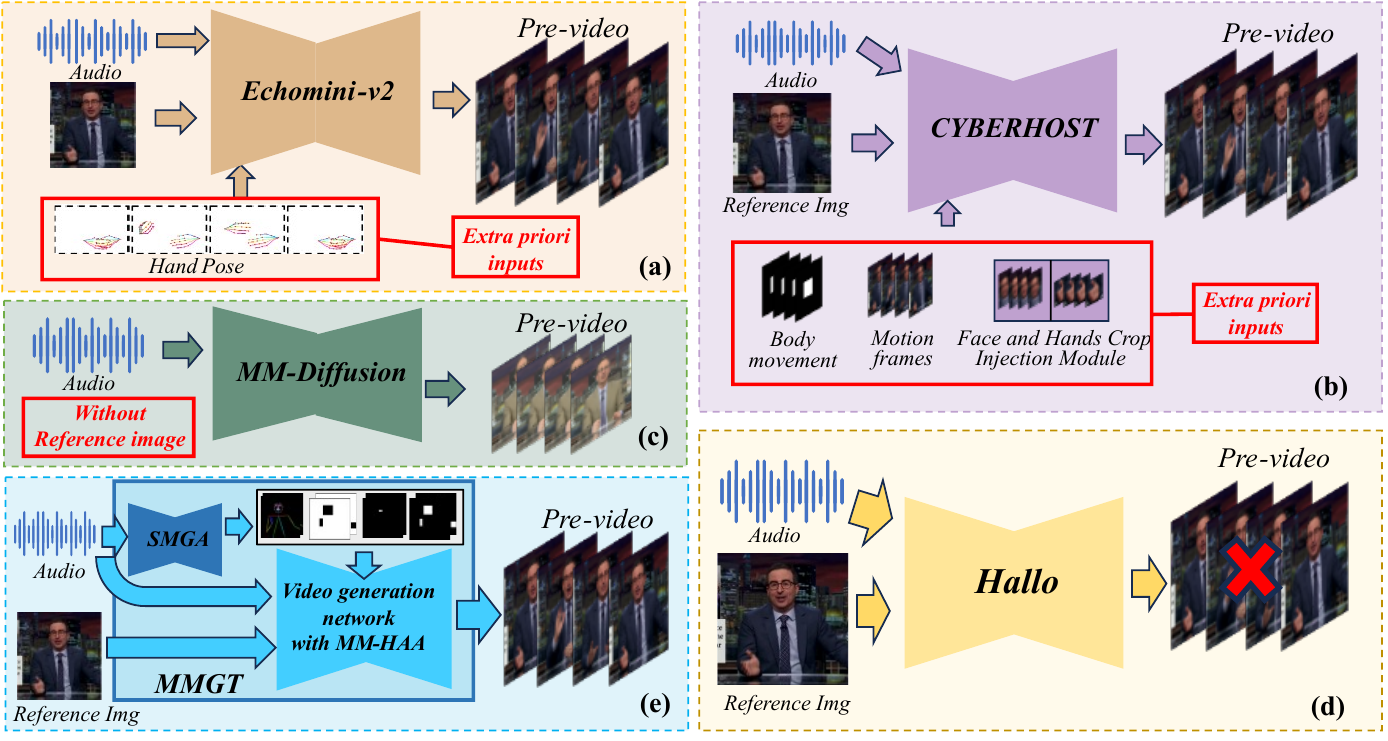}}%
\caption{Overview of existing co-speech gesture video generation models. In contrast to other methods~\cite{meng2024echomimicv2,ruan2022mmdiffusion,lin2024cyberhost, xu2024hallo}, our approach (MMGT) does not require additional a priori inputs in the inference process.}
\vspace{-2em} 
\label{fig:1}
\end{figure}

\section{Introduction}
\lettrine[lines=2, findent=2mm, nindent=0mm]{C}{o-speech} facial expression and gestures, as typical non-verbal behaviors, play a crucial role in human communication\cite{goldin1999role,burgoon1990nonverbal}. In a speech video, combining the speaker's head movements and gestures can make a person's speech more vivid and help convey the meaning more clearly. Therefore, generating videos of the speaker's gestures and head movements that are synchronized with the speech has attracted significant interest from researchers. Unified co-speech gesture and face video generation has great potential for virtual worlds, Digital Human Development\cite{10584563}, and multimedia applications\cite{Pataranutaporn2021AIgeneratedCF}. Although some current methods \cite{hogue2024diffted, lin2024cyberhost, meng2024echomimicv2, liu2024tango, ruan2022mmdiffusion,liu2022audio, he2024co} can generate vivid speech gesture videos, unfortunately, videos generated solely from audio and reference image often lack quality \cite{hogue2024diffted, he2024co}. On the other hand, some approaches \cite{meng2024echomimicv2, lin2024cyberhost} achieve better quality but require additional prior information beyond audio and inferred images. These prior elements are not as easily accessible as audio, which limits the practical use of the technology.

\begin{figure*}[t]
    \centering
    \includegraphics[width=1\textwidth]{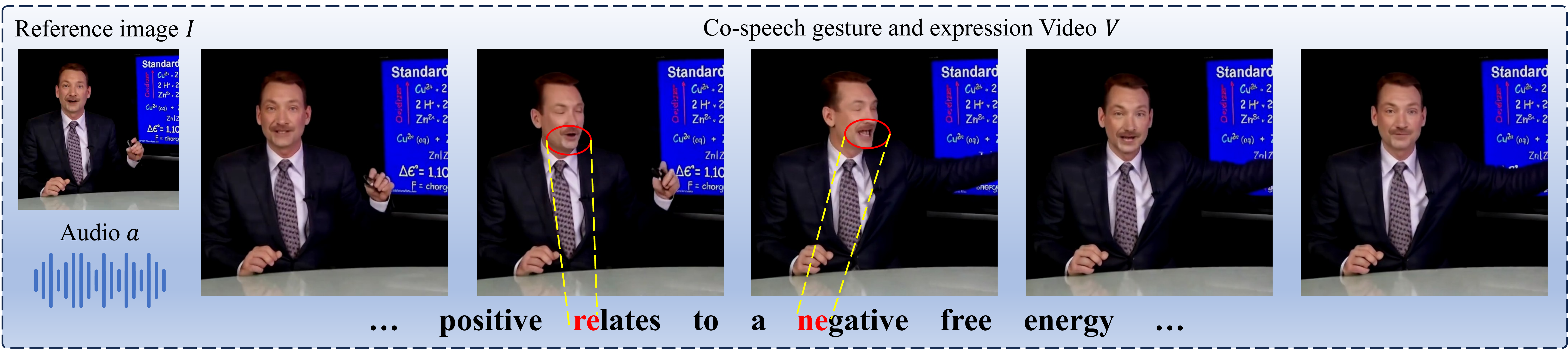} 
    \caption{\textbf{Examples of our generated co-speech gestures video.} The lips marked with red circles correspond to the bold red letters.}
    \vspace{-1.5em}
    \label{fig:2}
\end{figure*}

Existing methods mainly use end-to-end video generation frameworks. As shown in Fig.~\ref{fig:1}(c), MM-Diffusion \cite{ruan2022mmdiffusion} generates the entire video directly from audio using a diffusion model but does not include a reference image, which limits effective control over the ID information of the characters in the video\cite{he2024co}. Diffted\cite{hogue2024diffted} uses multi-stage audio processing to generate gesture videos, while EchoMimicV2\cite{meng2024echomimicv2} and CyberHost\cite{lin2024cyberhost} generate gesture videos by combining audio with prior information about hand pose videos or by motion frames and body movement map enhancement techniques to generate gesture and face videos. 

However, these methods still have several limitations. First, it is challenging to accurately synchronize facial movements and smooth gestures using only audio as a condition \cite{ruan2022mmdiffusion}. Second, most methods depend on additional high-quality prior inputs besides audio during inference (see Fig.~\ref{fig:1}(a)(b)), which limits practical application. Third, lacking region-specific constraints, as shown in Fig.~\ref{fig:1}(c), the generated videos tend to exhibit significant deformations and distortions in critical regions such as the face and gestures. For instance, large movements like head rotations can cause head distortions in the generated videos \cite{hogue2024diffted}.

To address these challenges, we propose the Motion Mask Guided Two-Stage Network (\textbf{MMGT}), which consists of two stages for video generation. In the first stage, we introduce the Spatial Mask Guided Audio2Pose Generation Network (\textbf{SMGA}) to generate a pose video guided by audio input. This pose videos not only captures synchronized facial expressions but also naturally incorporates hand gestures. Based on this, as shown in Algorithm~\ref{alg:draw_facepose_bbox}, we further compute the maximum bounding box for each region to obtain the motion mask.

In the second stage, we introduce a novel Motion Masked Hierarchical Audio Attention \textbf{(MM-HAA)} Module, which uses audio to dynamically enhance the details of specific body regions while pose sequences guide the motion of reference image. By adding this module, our framework can more precisely control the generation of simultaneous interpretation videos, as shown in Fig.~\ref{fig:2}, significantly boosting both realism and expressive fidelity.

As shown in Fig.~\ref{fig:1}, the comparison of input requirements and output capabilities across different methods is illustrated. Unlike EchoMimicV2\cite{meng2024echomimicv2} and CyberHost\cite{lin2024cyberhost}, which require additional inputs such as hand pose sequences or body movement modules, our proposed method, MMGT, achieves synchronized generation of gestures and lip movements using only audio and a single reference image as inputs, without relying on any extra prior information.
In summary, the main contributions of this work are as follows:

1. We propose the Spatial Mask Guided Audio2Pose Generation Network to generate a pose video and a motion mask from a speech signal and an initial pose. This integrated design provides natural and synchronized control of face and hand movements, leading to cohesive and realistic co-speech animations.

2. We introduce the Motion Masked Hierarchical Audio Attention Module for video generation. By integrating audio conditions with motion masks, the model adaptively focuses on specific body regions, such as hand gestures, lips, and face, enabling fine-grained detail enhancement.

3. Our MMGT network only needs audio and a single reference image to generate high-quality face and gesture videos, without extra prior inputs, significantly improving practicality.

Extensive experiments demonstrate that our framework significantly outperforms existing methods in generating vivid, realistic, speech-synchronized, and temporally stable gesture videos. By effectively capturing the fine-grained synchronization of face and gestures with speech, MMGT advances the state-of-the-art in co-speech video generation, delivering results that are qualitatively and quantitatively superior to existing techniques.

\begin{figure*}[t]
    \centering
    \includegraphics[width=1\textwidth]{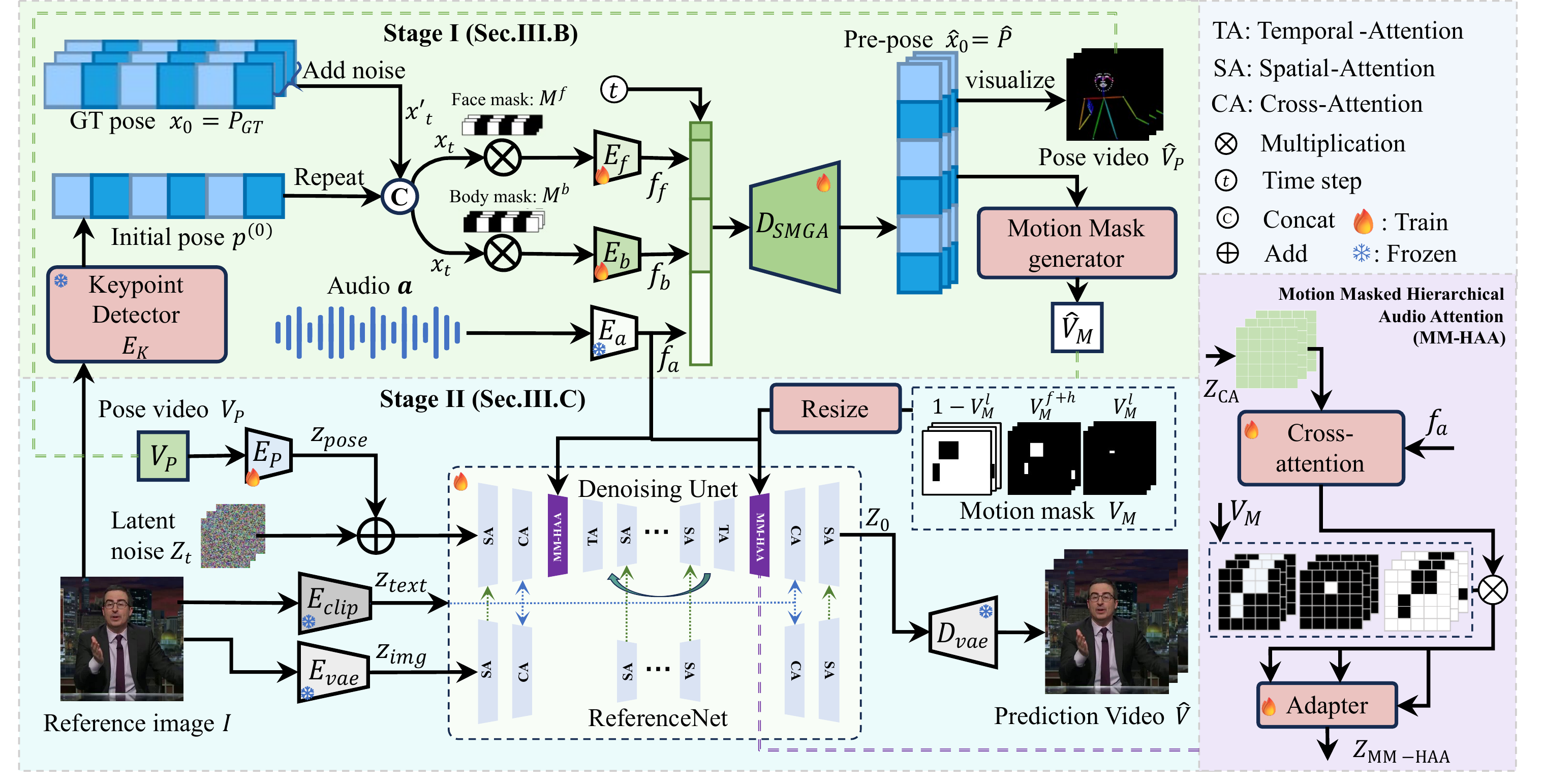}
    \vspace{-1.7em}
    \caption[width=0.8\textwidth]{\textbf{Overview of the Proposed MMGT Framework.} The framework operates in two stages: In Stage I, the SMGA network generates motion feature (pose video \( \hat{V}_P \), motion mask \( \hat{V}_M \)), based on the input audio \( a \) and initial pose \( p^{(0)} \). In Stage II, the Denoising UNet uses \( Z_{\text{pose}} \) and \( Z_{\text{text}} \), while ReferenceNet incorporates \( Z_{\text{pose}} \), \( Z_{\text{text}} \), and \( Z_{\text{img}} \) to produce the final predicted video \( \hat{V} \). On the right, the MM-HAA module enhances \( \hat{V}_M \) by aligning audio features \( f_a \) with cross-attention embeddings \( Z_{\text{CA}} \). The green double-dashed line indicates the inference process, where \( V_P \) and \( V_M \) replace \( \hat{V}_P \) and \( \hat{V}_M \) when transitioning from training to inference.}
    \vspace{-1.5em}
    \label{fig:3}
\end{figure*}

\section{RELATED WORK}
\subsection{Co-speech Face Video Generation}

In the field of co-speech face video generation, early studies \cite{kumar2017obamanet, prajwal2020lip, lipsgan} mainly concentrated on the lip region due to its strong link to audio. These works \cite{prajwal2020lip, lipsgan} leveraged GAN-based networks \cite{creswell2018generative} to synchronize lip movements in the video with the corresponding audio. 
Later, researchers expanded audio-lip synchronization to include synchronized generation of co-speech head movements, facial expressions, and emotions \cite{zhang2023sadtalker, tian2024emo, 10816597, 10237279}. Some studies \cite{wei2024aniportrait, xu2024hallo, chen2024echomimic, tian2024emo} utilized pre-trained SD-diffusion models \cite{rombach2022high} to generate high-resolution and dynamic talking head videos from reference image guided by audio. 

However, due to the limitations of current models, it is difficult to expand face-specific video generation tasks to full upper-body video creation simply by retraining with a different dataset, as shown in Fig.~\ref{fig:1}(d). These methods typically require the introduction of additional motion control features, otherwise, full-body control cannot be achieved. Furthermore, current datasets also have inherent limitations: LRS2, VoxCeleb, and MEAD \cite{afouras2018deep, nagrani2020voxceleb, wang2020mead} only cover face regions without including the full body, while the HDTF dataset \cite{zhang2021flow} includes the entire upper body but lacks significant audio-correlated dynamic gestures. These issues collectively hinder the adaptation of existing co-speech face video generation methods for gesture region synthesis. By designing a well-structured network and utilizing the PATS dataset\cite{ahuja2020style, ahuja-etal-2020-gestures, DBLP:journals/corr/abs-2007-12553}, our method effectively extends face video generation to co-speech upper-body video generation.

\begin{figure}[t]
    \centering
    \includegraphics[width=0.49\textwidth]{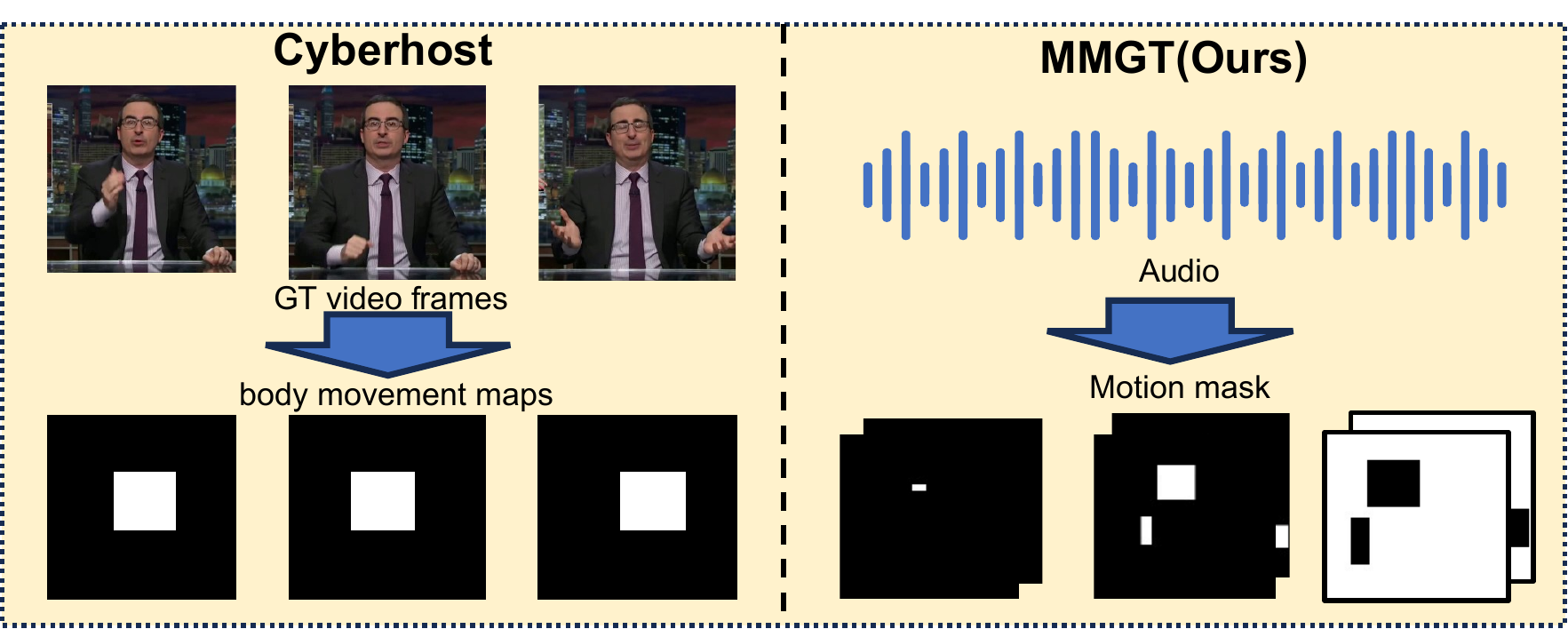} 
    \caption{Comparison of motion masks (MMGT) and body movement maps CyberHost\cite{lin2024cyberhost}.}
    \vspace{-1.5em}
    \label{fig:4}
\end{figure}

\subsection{Co-speech Gesture Generation}
In the field of gesture generation, early studies mainly relied on human skeleton data to generate lively gesture sequences driven by audio \cite{zhang2022motiondiffuse, wang2024detdiffusion, yang2023diffusestylegesture}. Recent research has incorporated multiple modalities based on audio, such as emotion \cite{chhatre2024emotional} and style \cite{yang2023diffusestylegesture}, to enhance the controllability of gesture sequence generation. Approaches like EMAGE \cite{liu2024emage} and DiffSHEG \cite{chen2024diffsheg} have expanded co-speech gesture generation to include head movement synthesis. However, these methods often ignore appearance information entirely \cite{he2024co}, leading to outputs that do not match users’ visual perception.

\subsection{Co-speech Gesture Video Generation}
Currently, gesture video generation methods are mainly divided into retrieval-based and generation-based methods. Zhou et al. \cite{zhou2022audio} were the first to frame gesture video generation as a task of reproducing video frames aligned with audio, generating high-quality gesture videos through a video frame retrieval method. TANGO \cite{liu2024tango} further advanced this approach by employing an action graph-based retrieval technique to generate co-speech body gesture videos. Although retrieval-based methods deliver good results with limited computational resources, they have significant limitations, such as the inability to generate new gestures and the lack of modeling the connection between the speaker’s face and audio.

ANGIE\cite{liu2022audio} was the first to clearly define the problem of co-speech gesture video generation, using unsupervised features MRAA\cite{siarohin2021motion} to simulate body movements. However, due to its linear nature, MRAA struggles to capture complex motion regions, thereby limiting the quality of generated videos. More recent methods, such as \cite{he2024co, hogue2024diffted}, adopt non-linear TPS transformations \cite{bookstein1989principal} to decouple human motion keypoints from images. These methods generate full keypoint sequences driven by audio, which are then used to produce compact optical flow for animating images and creating smooth gesture videos. 

Despite their advancements, generation-based approaches face common challenges: they fail to simultaneously generate detailed finger and lip movements, exhibit significant artifacts and deformation, and suffer from limited generalizability. CyberHost’s \cite{lin2024cyberhost} performance gains largely stem from the integration of extra a priori cues during both training and inference, namely the Body Movement Map and Motion Frames shown in Fig.~\ref{fig:1}(a) and (b). EchoMimicV2 \cite{meng2024echomimicv2}, meanwhile, synthesises gesture videos by conditioning on the input audio and hand-pose videos, then applies dedicated enhancement modules to generate co-speech videos. As shown in Fig.~\ref{fig:4}, CyberHost's \cite{lin2024cyberhost} input body movements maps are built from ground videos, which can only capture large-scale body movements. In contrast, our motion masks are derived from audio, which both accurately represent motion and enrich the detail of the corresponding regions.


\section{Proposed Method}
We propose the \textbf{MMGT} network, as shown in Fig.~\ref{fig:3}, which ensures synchronized control of both face and hand gestures while also compensating for regional details. 
Given a segment of audio denoted it as \( \mathbf{a} = [a_{0}, \dots, a_{N}] \) and a reference image \( I_{0} \in \mathbb{R}^{3 \times W \times H} \), our framework generates a co-speech video with \( N \) frames, called \( \hat{V} = [I_{0}, \hat{I}_{1} \dots, \hat{I}_{N-1}] \in \mathbb{R}^{N \times 3 \times W \times H} \). The process of co-speech video generation is divided into two stages: motion mask and pose videos generation (Stage I) and motion-driven and detail refinement (Stage II). Therefore, our inference pipeline can be formulated as
\begin{align}
&\hat{V}_P, \hat{V}_M = SMGA(E_k(I_0), a), \\
&\hat{V} = \mathcal{G}_v(I_0, \hat{V}_P, \hat{V}_M, a),
\end{align}
where \( E_k(\cdot) \) represents the initial pose extractor, responsible for obtaining the initial pose \( p^{(0)} \) from the reference image \( I_0 \). In the first stage, our proposed \( SMGA(\cdot) \) network, built upon the DiT architecture \cite{dit}, takes audio \( a \) as a conditioning signal to generate the complete pose sequence \( \hat{x}_0 \). In the second stage, the video generation model \( \mathcal{G}_v(\cdot) \) incorporates an reference-net and a temporal module based on the SD architecture \cite{nokey}. It leverages the motion mask \( \hat{V}_M \) generated in the first stage, along with the pose video \( \hat{V}_P \) and audio \(\ a \), as conditioning inputs to synthesize the final video \( \hat{V} \).

\subsection{Overall Framework Architecture}

In the first stage, the SMGA network aims to capture the temporal and semantic relationships between the pose sequence and the audio input. As shown in Fig.~\ref{fig:5}, it includes several key components. The motion block captures the correlation between audio and posture to obtain face and overall body features, and the merge block combines the two to generate the final output.

In the second stage, our framework leverages a Denoising UNet with three attention modules to refine motion details by integrating motion masks, pose videos, audio, and spatial cues from a reference image. Among these modules, the Spatial Attention (SA) and Temporal Attention (TA) are directly adopted from Animate-Anyone\cite{hu2024animate} to enhance texture details and ensure smooth transitions. Our key contribution lies in the Motion Masked Hierarchical Audio Attention (MM-HAA) module, which is inspired by the Hierarchical Audio Attention in Hallo \cite{xu2024hallo} but innovatively replaces the static global mask with a dynamic motion mask derived from the pose sequence, as shown in Fig.~\ref{fig:11}. This design enables audio features to focus on actively moving regions in each frame, leading to more precise audiovisual alignment and improved temporal-spatial consistency in the generated video.

\subsection{Pose Video and Motion Mask Generation}
As shown in Fig.~\ref{fig:3}, the main goal of the first stage is to synthesize pose videos and motion masks driven by audio inputs. To do this, we developed a Spatial Mask-Guided Audio2Pose Generation (SMGA) Network, which generates both the complete pose video \( \hat{V}_P \) and the corresponding motion mask \( \hat{V}_M \). Among them, the pose video \( \hat{V}_P \) effectively captures the core dynamics of motion, while the motion mask \( \hat{V}_M \) emphasizes key parts of significant movement, such as hand gestures and face, thereby enhancing the accuracy of motion representation.

\begin{figure}[t]
\centerline{%
\includegraphics[width=0.4\textwidth]{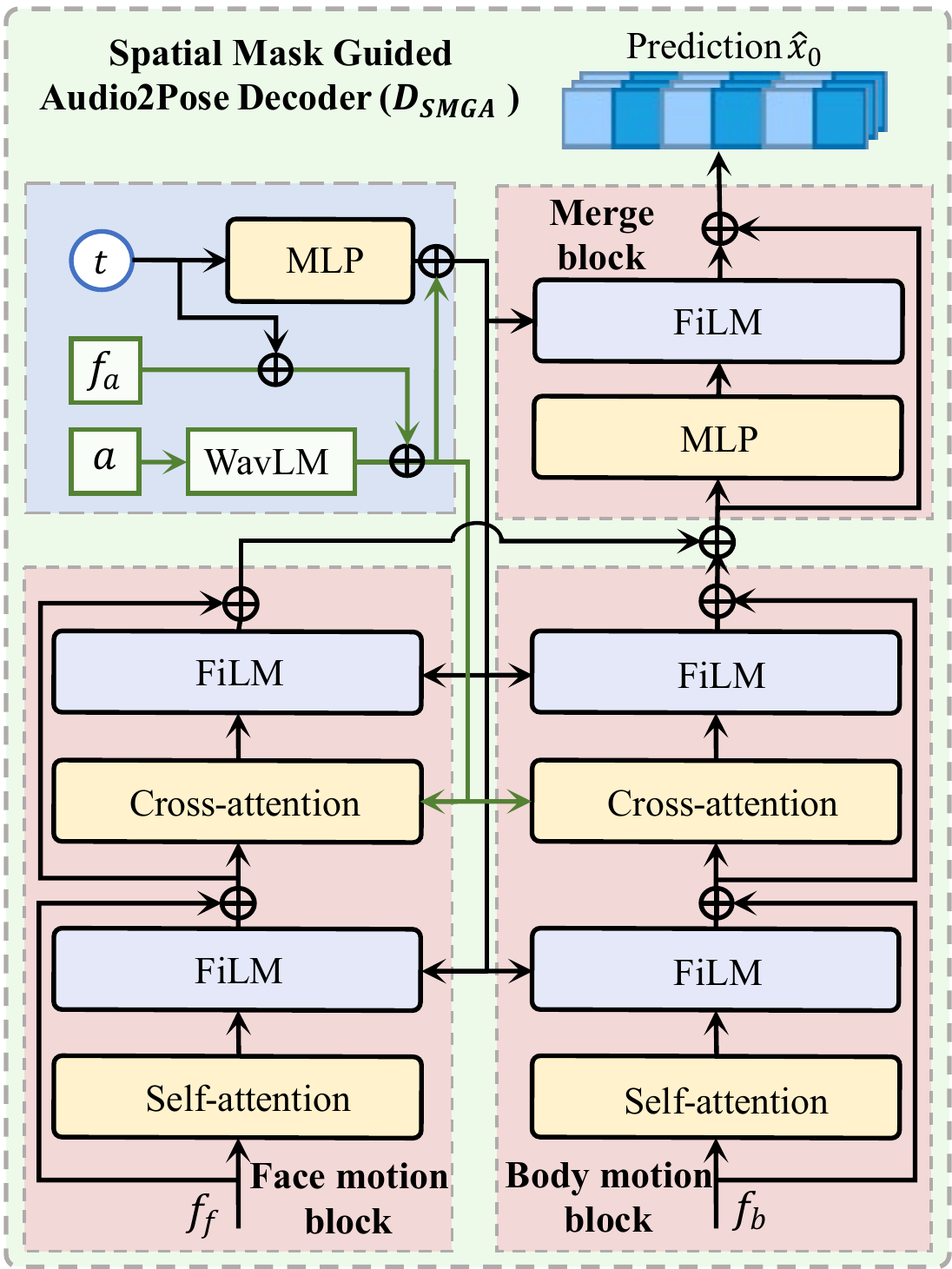}}%

\vspace{-0.5em} 
\caption{Architecture of the $D_{SMGA}$.}
\vspace{-1.5em}
\label{fig:5}
\end{figure}

\textbf{Spatial Mask-guided Pose Video Generation.}
The initial pose \( p^{(0)} \) is extracted from the reference image \( I_0 \) using a pre-trained keypoint detector\cite{yang2023effective} \( E_K \),
\begin{align}\label{eq:6}
p^{(0)} = E_K(I_0),
\end{align}
 To introduce variations for motion modeling, Gaussian noise is added to the ground-truth pose \( x_0 = P_{\text{GT}} \in \mathbb{R}^{T \times C_p \times 3} \) to obtain noisy motion representations,
\begin{align}\label{eq:7}
x'_t = x_0 + \mathcal{N}(0, \sigma^2),
\end{align}
where \( x'_t \) represents the noisy pose at time step \( t \). The initial pose \( p^{(0)} \) is repeated along the temporal dimension and added to \( x'_t \) obtain the input pose \( x_t \),
\begin{align}\label{eq:8}
x_t = x'_t + \mathcal{R}(p^{(0)}),
\end{align}
where \(\mathcal{R}(\cdot)\) denotes the replication of N copies of the delayed time dimension.
The input pose \( x_t \) is categorized into face spatial \( C_f \)  and body spatial \( C_b \) as
\begin{align}\label{eq:1}
x_t = [p^1, p^2, \ldots, p^N] \in \mathbb{R}^{N \times (C_{f}+C_{b}) \times 3}, 
\end{align}
To apply masking to \( x_t \) along the feature dimension \( C \), we create the face spatial mask \( M^f \) as
\begin{align}\label{eq:9}
M^f = 
\begin{cases} 
1, & \text{if } C \in C_f, \\ 
0, & \text{otherwise}.
\end{cases}
\end{align}

\begin{figure}[t]
    \centering
    \includegraphics[width=0.49\textwidth]{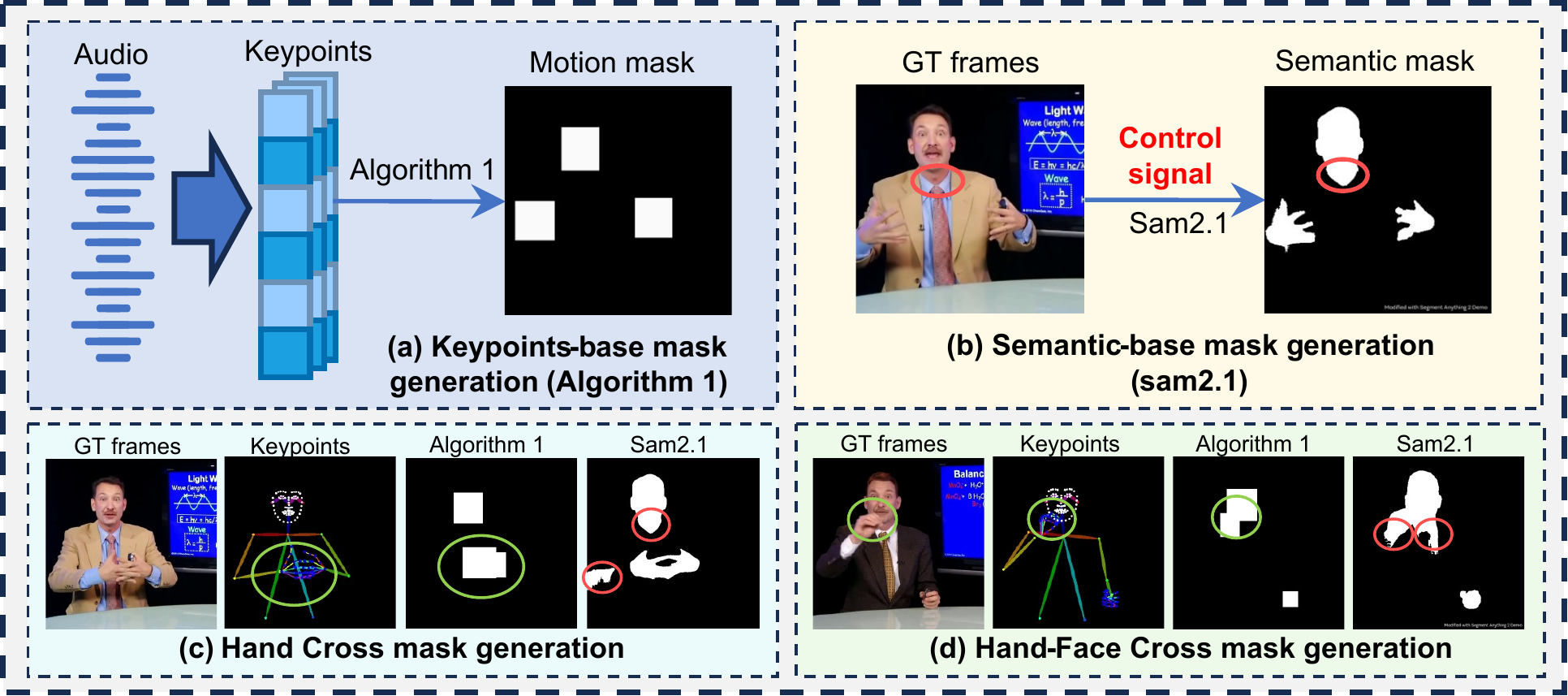} 
    \caption{\textbf{Comparison of Algorithm 1 and Semantic Segmentation for Mask Generation.} We further show a comparison of the stability of different mask generation methods when body parts overlap.}
    \vspace{-1.5em}
    \label{fig:6}
\end{figure}

Next, the complementary body spatial mask \( M^b \) is obtained as \( M^b = 1 - M^f \). These spatial mask are then applied to the input \( x_t \) to isolate the respective feature representations, as defined below 
\begin{align}\label{eq:10}  
x_t^f = x_t \odot M^f, \quad x_t^b = x_t \odot M^b,  
\end{align}  
where \( x_t^f \) and \( x_t^b \) correspond to the spatial features for head and body pose, respectively. Distinct encoders are employed to process these input conditions, transforming them into separate feature embeddings
\begin{align}\label{eq:11}
\begin{cases} 
f_a = E_a(\mathbf{a}), \\
f_f = E_f(x_t^f), \\
f_b = E_b(x_t^b),
\end{cases}
\end{align}
where \( f_a \) represents audio feature embedding, with \( E_a \) being a pre-trained audio encoder \cite{he2024co}, while \( E_f \) and \( E_b \) are encoders for the face and body pose features, \( f_f \) and \( f_b \), respectively. Finally, the output of the first stage is expressed as
\begin{align}\label{eq:12}
\hat{x}_0 = D_{\text{SMGA}}(f_f, f_b, f_a, t),
\end{align}
where \( t \) represents the temporal embedding. Subsequently, these features are fed into the Spatial Mask-Guided Audio2Pose (SMGA) Decoder, denoted as \( D_{\text{SMGA}} \), which decodes them to produce a sequence of overlapping speech poses \( \hat{x}_0 \). The generated pose sequence \( \hat{x}_0 \) is then visualized as the pose video \( V_P \).

\textbf{Motion Mask Generation.}
As shown in Algorithm~\ref{alg:draw_facepose_bbox}, we introduces a new way to create motion masks based on pose sequence, represented as \(P \in \mathbb{R}^{T \times C_p \times 3}\). Here, \(C_p = C_f + C_h + C_l\) corresponds to face, hand gestures, and lips, respectively. In each frame of a video sequence, our algorithm transforms normalized keypoint coordinates into pixel coordinates and calculates bounding boxes for specified regions. These bounding boxes are then utilized to construct binary masks, setting active regions to a value of 255. The generated masks include \(V^f_M\) for face, \(V_M^l\) for limb movements, and \(V_M^h\) for hand gestures. Notably, as the head and hand masks are non-overlapping, they are combined into a single mask, denoted as \(V_M^{f+h}\), and we set the background mask to \(V_M^{b} = (255 - V_M^{f+h})\). In the inference process of our framework, the generation process of the motion mask is shown in Fig.~\ref{fig:6}(a). Even in cases of limb occlusion or keypoints overlap, as shown in the green circle in Fig.~\ref{fig:6}(c)(d), the bounding boxes generated from different keypoints will only create overlapping areas instead of gaps, thus naturally ensuring full coverage of the motion region. It is superior to SAM2.1\cite{ravi2024sam2} in terms of stability of mask generation and is better suited to our framework.

\begin{figure}[t]
    \renewcommand{\algorithmicrequire}{\textbf{Input:}}
    \renewcommand{\algorithmicensure}{\textbf{Output:}}
    \vspace{-1em}
    \begin{algorithm}[H]
        \caption{Motion Mask Generator $G_M(\cdot)$}
        \begin{algorithmic}[1]
            \REQUIRE Pre-pose $\hat{P}\in \mathbb{R}^{N \times {C_p}}, {C_p}=(C_f+C_h+C_l)$
            \ENSURE Masks videos $V_M^{f}, V_M^{l}, V_M^{h}\leftarrow V_M \in \mathbb{R}^{N \times H \times W}$
            \STATE Initialization $V_M\leftarrow \mathbf{0}$
            \FOR{each frame $t = 1$ to $N$}
                \STATE $bbox \leftarrow \text{None}$
                \STATE $p_i \leftarrow \hat{P}[t]$
                \STATE Initialize $min_x, min_y \leftarrow W, H$, $max_x, max_y \leftarrow 0, 0$
                \FOR{each $p_{i} \in p_i$}
                    \STATE $x, y \leftarrow \text{int}(p_{i} \cdot [W, H])$
                    \IF{$x > 0$ and $y > 0$}
                        \STATE Update $min_x, min_y, max_x, max_y$
                    \ENDIF
                \ENDFOR
                \IF{$min_x < max_x$ and $min_y < max_y$}
                    \STATE $V_M[t, min_y:max_y, min_x:max_x] \leftarrow 255$
                \ENDIF
            \ENDFOR
            \STATE \textbf{return} $V_M^{f}, V_M^{l}, V_M^{h}$
        \end{algorithmic}
        \label{alg:draw_facepose_bbox}
    \end{algorithm}
    \vspace{-2.5em}
\end{figure}

\textbf{SMGA Decoder Structure.}
The \(D_{SMGA}\) is carefully designed to capture the temporal and semantic relationships between pose sequence and audio input.  As shown in Fig.~\ref{fig:5}, its architecture consists of two motion blocks and a merging module. The two motion blocks process the input features \(f_f\) or \(f_b\) that represent different body parts separately through a combination of self-attention, FiLM\cite{perez2018film, tseng2023edge}, and cross-attention mechanisms to capture the correlation between different body parts feature $f$ and audio feature \(f_a\). Among them, the self-attention mechanism is used to capture the correlation of the pose sequence itself, and the calculation formula is as
\begin{align}\label{eq:13}
\text{Att}(Q_f, K_f, V_f) = \text{softmax}\left(\frac{Q_fK_f^\top}{\sqrt{d_k}}\right)V_f,
\end{align}
Then, the cross-attention mechanism in different motion modules will capture the different correlations between different body parts $f$ and audio feature \(f_a\). The specific formulas are as
\begin{align}\label{eq:14}
\text{Att}(Q_f, K_{f_a}, V_{f_a}) = \text{Softmax}\left(\frac{Q_fK_{f_a}^\top}{\sqrt{d_k}}\right)V_{f_a},
\end{align}
where \( Q_f \) is derived from pose sequence (\( f_f \) or \( f_b \)), and \( K_{f_a} \), \( V_{f_a} \) are derived from audio features \( f_a \). Through this distraction mechanism, we achieve coherent and coordinated speech actions for different body regions. 

Next, inspired by EDGE \cite{tseng2023edge}, the joint representation is modulated by a FiLM block \cite{perez2018film}. FiLM provides only affine modulation in the audio condition to enhance model representation. The diffusion time-step embedding $t$ with the audio feature $f_a$ after a linear layer MLP:
\begin{align}
     x_i = f_a + \mathrm{MLP}(t), 
\end{align}
The FiLM block predicts the feature-wise scaling $\gamma_{i,c}$ and bias terms $\beta_{i,c}$:
\begin{align} 
    \gamma_{i,c}=f_c(x_i),\quad
    \beta_{i,c}=h_c(x_i),
\end{align}
where $f_c(\cdot)$ and $h_c(\cdot)$ are learnable functions. FiLM then applies channel-wise affine transformations to three motion branches, $F_{i,c}$ with $c\in\{\text{face},\text{body},\text{merge}\}$:
\begin{align}
    \operatorname{FiLM}(F_{i,c}) = \gamma_{i,c}\,F_{i,c}+\beta_{i,c},
\end{align}
where Face block injects $\gamma_{i,c}, \beta_{i,c}$ to drive audio-synchronized face. Body block injects $\gamma_{i,c}, \beta_{i,c}$ to synchronize gesture rhythms with the soundtrack. Merge block concatenates face and body features at the decoder’s end and applies a final FiLM gating.

\begin{figure}[t]
\centerline{\includegraphics[width=0.48\textwidth]{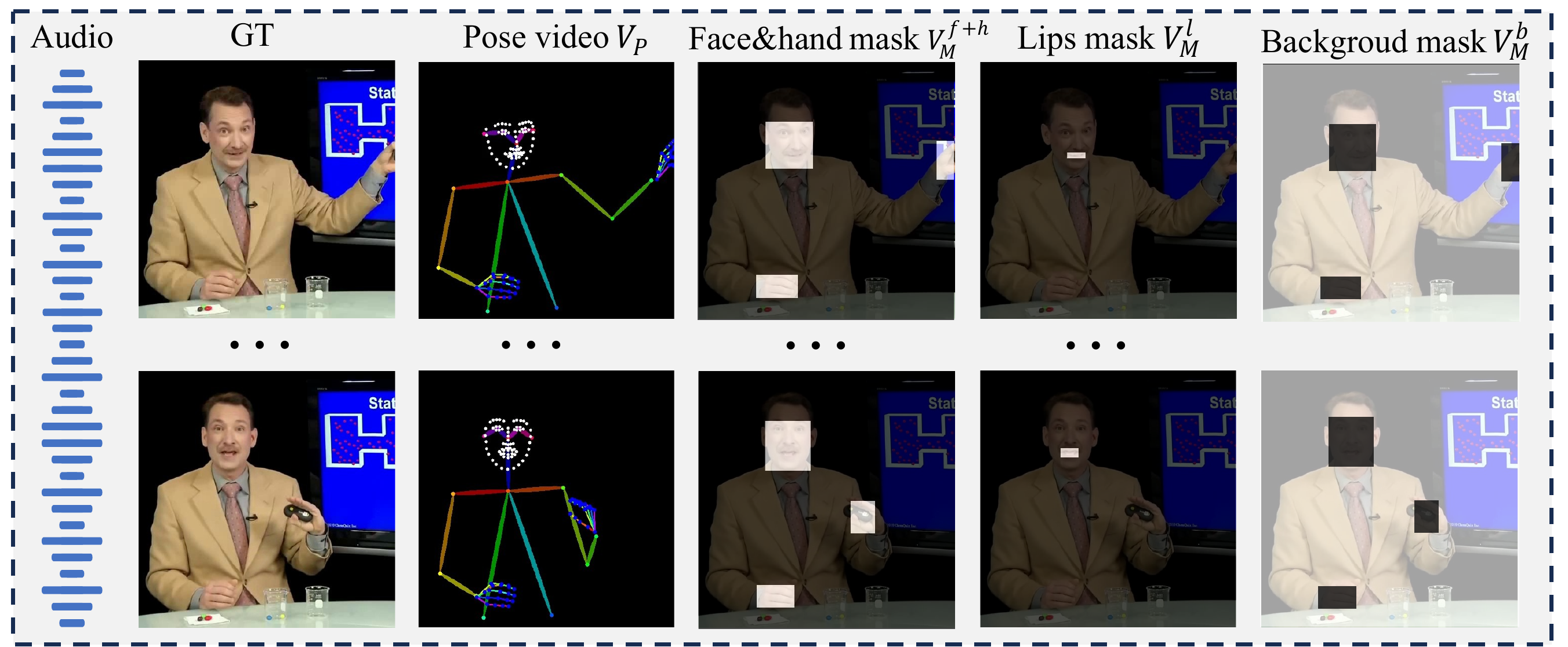}}
\vspace{-0.5em}
\caption{The details of inputs during model training.}
\vspace{-1.5em}
\label{fig:7}
\end{figure}

\textbf{Loss Function.} The training process optimizes the SMGA network using a multi-component loss function to ensure accurate pose generation. The reconstruction loss is defined as
\begin{align}\label{eq:16}
\mathcal{L}_{\text{rec}} = \frac{1}{T} \sum_{t=1}^T \| x_t - \hat{x}_t \|_2^2,
\end{align}
where \( x_t \) and \( \hat{x}_t \) represent the ground-truth and predicted pose sequence, respectively.
The velocity loss is
\begin{align}\label{eq:17}
\mathcal{L}_{\text{vel}} = \frac{1}{T-1} \sum_{t=1}^{T-1} \| (x_{t+1} - x_t) - (\hat{x}_{t+1} - \hat{x}_t) \|_2^2,
\end{align}
The acceleration loss is expressed as
\begin{align}\label{eq:18}
\mathcal{L}_{\text{acc}} &= \frac{1}{T-2} \sum_{t=1}^{T-2} \Big\| \left( x_{t+2} - 2x_{t+1} + x_t \right) \nonumber \\
&\quad - \left( \hat{x}_{t+2} - 2\hat{x}_{t+1} + \hat{x}_t \right) \Big\|_2^2,
\end{align}
Each loss term for head and body pose can be defined as
\begin{align}\label{eq:19}
\mathcal{L}_f = \mathcal{L}_{\text{rec}}^f + \mathcal{L}_{\text{vel}}^f + \mathcal{L}_{\text{acc}}^f, \quad \mathcal{L}_b = \mathcal{L}_{\text{rec}}^b + \mathcal{L}_{\text{vel}}^b + \mathcal{L}_{\text{acc}}^b,
\end{align}
where \( \mathcal{L}_f \) and \( \mathcal{L}_b \) correspond to the losses for head and body pose, respectively. Further details of the reconstruction, velocity, and acceleration terms can be expanded based on the specific metrics used for evaluation. The total loss is computed as
\begin{align}\label{eq:21}
\mathcal{L}_{\text{SMGA}} = \lambda_f \mathcal{L}_f + \lambda_b \mathcal{L}_b,
\end{align}
through sensitivity analysis, as shown in Table.~\ref{Tab:loss}, we selected \( \lambda_f = 3 \) and \( \lambda_b = 1 \) are weighting factors for the head and body pose losses. These weights control the relative contribution of each pose type to the overall optimization process.

\subsection{Motion-Driven Detail Refinement Video Generate}\label{Video_Generate}

\textbf{Conditional Input Preparation.}
The video generation process begins with preparing the conditional inputs. As shown in Fig.~\ref{fig:7}, the training inputs consist of audio signal \( a \), the original video \( V \), the pose video \( V_P \), and three types of motion masks: \( V^{f+h}_M \), \( V_M^l \), and \( V_M^{b} \). These inputs collectively provide the necessary information to guide the generation of the final video output. We extract a 12-frame video clip from the training set to represent a single step of the model, with the first frame of the clip designated as the reference image \( I_0 \). This reference image provides the foundational visual context for subsequent processing and generation within the model.

The CLIP \cite{radford2021learning} image encoder \( E_{\text{clip}} \) extracts semantic features $Z_{text}$ from the text associated with the reference image \( I_0 \), while the autoencoder \( E_{\text{vae}} \) \cite{kingma2013auto, van2017neural} encodes the implicit visual representation of the image. For audio input \( a \), we utilize \( E_a \) \cite{baevski2020wav2vec, chen2022wavlm} to extract the corresponding audio features \( f_a \). For the pose video \( V_P \), a specially designed pose encoder \cite{hu2024animate} is employed to extract pose features, denoted as \( Z_{\text{pose}} \). Additionally, motion masks \( V_M = \{V_M^{f+h}, V_M^l, V_M^{b}\} \) of different sizes are generated using Gaussian blurring and resizing operations. These masks are then fed into different layers of the model. All extracted features are temporally aligned to ensure the generated video is synchronized with the input audio.

\begin{figure*}[t]
    \centering
    \includegraphics[width=0.99\textwidth]{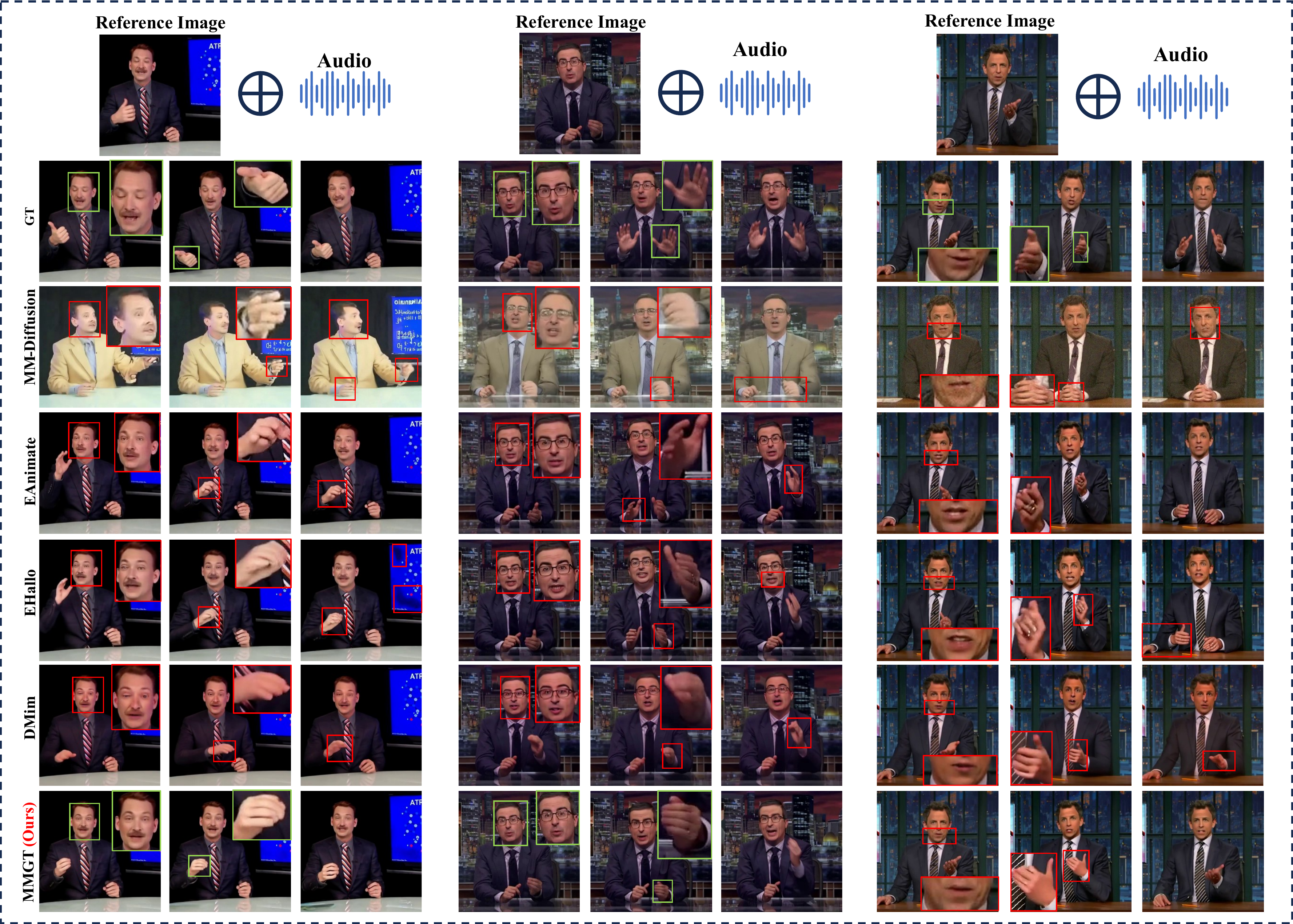}
    \vspace{-0.5em}
    \caption[width=0.11\textwidth]{\textbf{Qualitative Comparison of Upper Body Co-speech Video Generation Models on the PATS Dataset.} This figure compares the performance of different co-speech video generation models (MM-Diffusion\cite{ruan2022mmdiffusion}, EDGE + Animate-anyone (EDGE + Animate) \cite{tseng2023edge, hu2024animate}, EDGE + Hallo\cite{tseng2023edge, xu2024hallo}, DiffGest + MimicMo\cite{zhu2023taming,zhang2024mimicmotion} and Ours MMGT) in generating realistic upper-body movements, speech synchronization, and maintaining video quality}.
    \vspace{-2.5em}
    \label{fig:8}
\end{figure*}

\textbf{Motion Masked Hierarchical Audio Attention.}
The Motion Mask-based Hierarchical Audio Attention (MM-HAA) module is a vital part of the second stage of our framework, designed to refine motion features by dynamically aligning spatial details, motion masks, and audio features. Unlike fixed-mask methods used in previous works \cite{xu2024hallo}, MM-HAA uses motion masks, \(V_M^{f+h}\), \(V_M^l\), and \(V_M^b\), which are derived from the pose sequence generated in the first stage. These masks adaptively focus attention on specific regions, with \(V_M^{f+h}\) focusing on face and hand gestures,  \(V_M^l\) focusing on lip movements, and \(V_M^b\) focusing on the background area. Through a cross-attention mechanism, MM-HAA aligns intermediate hidden states \(Z_{CA}\) with specific temporal and frequency audio features \(f_a\), ensuring spatial-temporal synchronization. The main operation in MM-HAA is defined as
\begin{align}\label{eq:23}
\text{Att}(Q_{Z_{CA}}, K_{f_a}, V_{f_a}) = \text{Softmax}\left(\frac{Q_{Z_{CA}}K_{f_a}^\top}{\sqrt{d_k}}\right)V_{f_a},
\end{align}
where \(Q_{Z_{CA}} = W_q Z_{CA}\), \(K_{f_a} = W_k f_a\), and \(V_{f_a} = W_v f_a\) represent the query, key, and value matrices, respectively, and \(d_k\) indicates the stability factor.
MM-HAA also combines these aligned features with motion masks to improve region-specific refinement. The masked hidden states are computed as
\begin{align}\label{eq:24}
    Z_{f+h}' &= Z_{CA} \odot V_M^{f+h}, \nonumber \\
    Z_l'     &= Z_{CA} \odot V_M^l, \nonumber \\
    Z_b'     &= Z_{CA} \odot V_M^b.
\end{align}
where \(\odot\) denotes element-wise multiplication. These masked representations are processed through convolution-based adapter modules, which utilize residual connections and convolutional layers to enhance feature richness
\begin{align}\label{eq:25}
Z_{\text{MM-HAA}} = \text{Adapter}_{f+h}(Z_{f+h}') &+ \text{Adapter}_{l}(Z_l') \nonumber \\
&+\text{Adapter}_{b}(Z_b'),
\end{align}
where \(\text{Adapter}_{f+h}\), \(\text{Adapter}_l\), and \(\text{Adapter}_b\) are the convolution-based refinement modules for \(V_M^{f+h}\), \(V_M^l\), and \(V_M^b\) respectively. 

This combined output \(Z_{\text{MM-HAA}}\) is fed into the Temporal Attention (TA) layers in the Denoising UNet, enabling further temporal refinement.
The hierarchical design of MM-HAA ensures accurate integration of motion masks, audio features, and spatial details across frames. By combining cross-attention with mask-based refinement, the module aligns audio-driven dynamics with spatially localized regions, effectively addressing the limitations of fixed-mask methods. Experimental results show that MM-HAA significantly enhances realism, synchronization, and expressiveness in generated co-speech videos. Its contributions to temporal coherence and region-specific enhancement are crucial to the overall success of the proposed framework.

\begin{figure*}[t]
    \centering
    \includegraphics[width=1\textwidth]{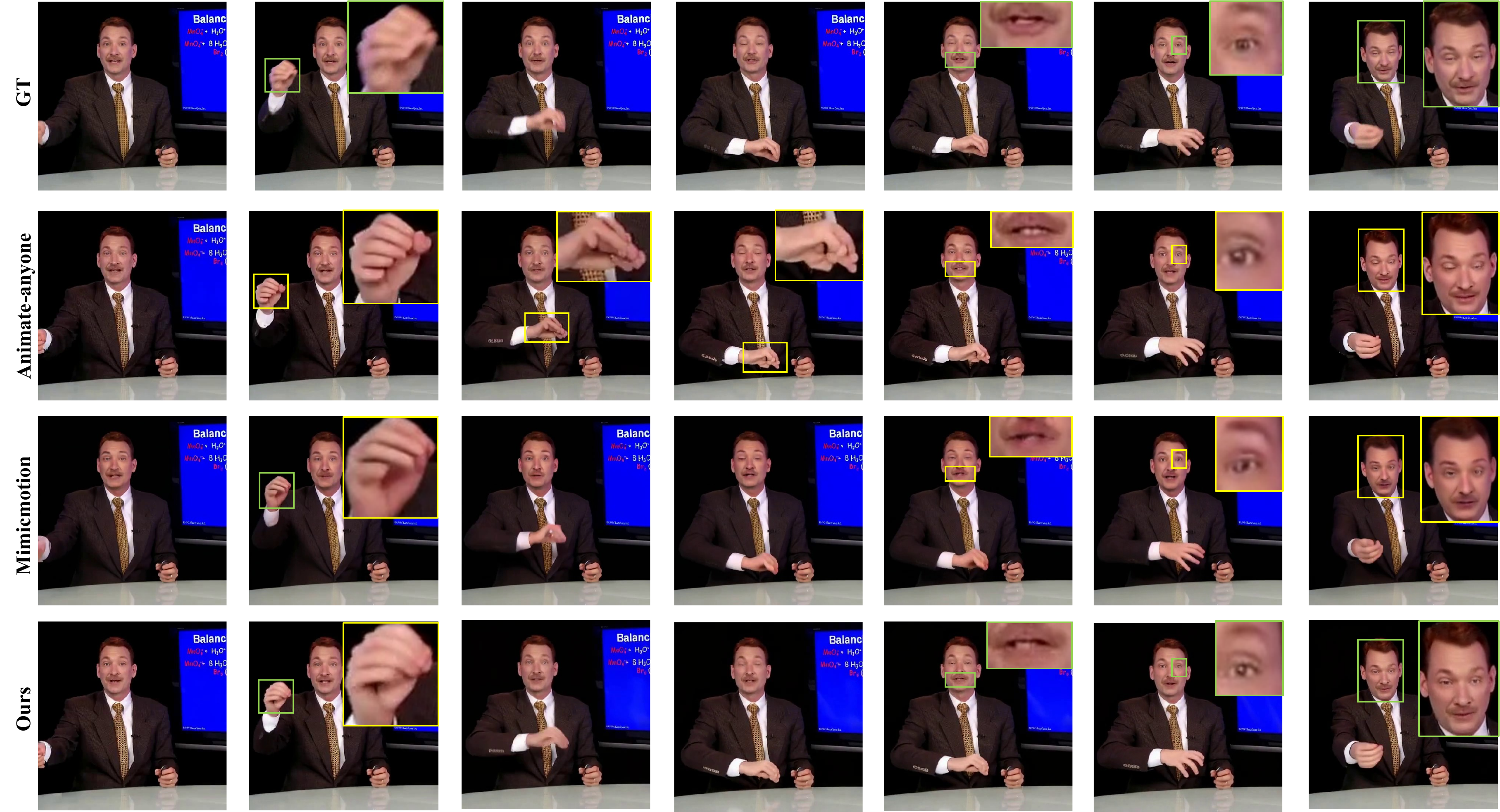}
    \vspace{-1.5em}
    \caption[width=1\textwidth]{\textbf{Qualitative Comparison of Pose-driven Video Generation.} This figure compares videos generated by  Animate-anyone \cite{hu2024animate}, MimicMo \cite{zhang2024mimicmotion}and MMGT(Ours) under the same pose-driven task.}
    \label{fig:9}
\end{figure*}

\begin{table*}[ht]
  \centering
  \vspace{-0.5em}
  \caption{Comparison of computational overhead of two training processes in the second stage of different methods.}
  \vspace{-0.5em}
  \resizebox{0.98\textwidth}{!}{
  \begin{tabular}{lcccccc}
    \toprule
    \textbf{Model (Processes I/Processes II)} &
    \textbf{ Params (GB)} &
    \textbf{Train GPU-hours (h)} &
    \textbf{Peak VRAM (GB)} &
    \textbf{Inference(s/per-Clip)}  &
    \textbf{Batch size-per GPU} & \textbf{Step} \\
    \midrule
    Animate-anyone\cite{hu2024animate}  & 6.3/1.69 & 7/13 & 41.4/35.88  &  151 & 2/1 & 29850/32500 \\
    MimicMo\cite{zhang2024mimicmotion} & 5.68 & 28 & 47.34  & 133 & 1 & 32500 \\
    Hallo\cite{xu2024hallo}  & 6.3/2.80 & 7/28 & 41.4/16.72 & 240 & 2/1 & 29850/32500\\
    MMGT(Ours) & 6.3/2.17 & 7/15 & 41.4/41.27 & 224 & 2/1 & 29850/32500 \\ 
    \bottomrule
  \end{tabular}}
  \label{table:cpmputation}
  \vspace{-2em}
\end{table*}

\begin{figure*}[t]
\centerline{
\includegraphics[width=1\textwidth]{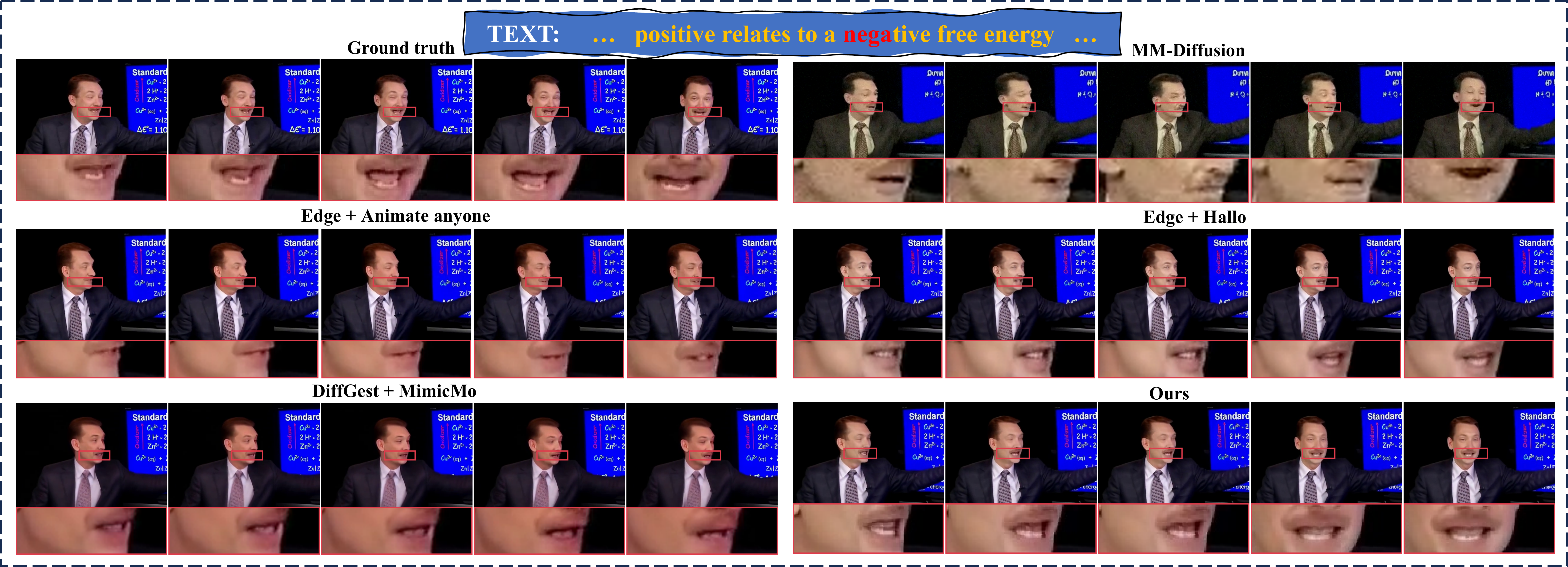}}
\vspace{-0.5em}
\caption{\textbf{Lips Synchronization with Audio.} A comparison of lip synchronization with audio across Ground Truth, MM-Diffusion\cite{ruan2022mmdiffusion}, EDGE + Animate-anyone \cite{tseng2023edge, hu2024animate}, EDGE + Hallo\cite{tseng2023edge, xu2024hallo}, DiffGest + MimicMo\cite{zhu2023taming, zhang2024mimicmotion} and our MMGT model.}
\label{fig:10}
\end{figure*}

\textbf{Loss Function.}
The model is optimized using a diffusion-based reconstruction loss
\begin{align}\label{eq:26}
\mathcal{L}^{latent}_{\text{rec}} = \frac{1}{T} \sum_{t=1}^T \| Z_t - \hat{Z}_t \|_2^2,
\end{align}
where \( T \) represents the number of time steps, \( Z_t \) is the generated latent representation at time step \( t \), and \( \hat{Z}_t \) is the corresponding target latent representation. The final video \( \hat{V} \) is reconstructed by decoding the refined latent representations \( Z_t \) through the decoder \( D_{\text{vae}} \), which transforms the latent space back into actual video frames.
\begin{figure}[t]
    \renewcommand{\algorithmicrequire}{\textbf{Input:}}
    \renewcommand{\algorithmicensure}{\textbf{Output:}}
    \vspace{-1em}
    \begin{algorithm}[H]
        \caption{Training Algorithm}
        \label{alg2}
        \begin{algorithmic}[1]
            \REQUIRE Reference image $I_0$, Driven audio $a$, and GT-video $V$, $\theta_1, \theta_2$ are parameters of $D_{SMGA}, D_{LV}$
            \ENSURE The optimal parameters $\theta_1^*, \theta_2^*$
            \STATE \textbf{Stage 1 Input Preparation}
            \STATE $x_{0} \leftarrow E_K(V) = \{p^0,...,p^L\} \triangleright \mathbb{R}^{L \times C}$
            \STATE $x_{r} \leftarrow repeat((E_K(I_0))) = \{p^0,...,p^0\}\triangleright \mathbb{R}^{L \times C}$
            \STATE $m_{body} = [0,1]\in\mathbb{R}^{L \times C_b}$
            \STATE $M = \{m_{body}, 1 - m_{body}\}$
            \STATE $x_t \leftarrow concat(x_0 + \mathcal{N}(0, \sigma^2), x_r)$ 
            \STATE $\mathbf{c}_{1} = \{M, x_{t}, E_a(a)\}$

            \STATE \textbf{Stage 2 Input Preparation}
            \STATE $z_{0} \leftarrow E_{vae}(V) \triangleright \mathbb{R}^{L \times C \times H\times W}$
            \STATE $z_t \leftarrow z_0 + \mathcal{N}(0, \sigma^2)$
            \STATE $V_M = G_M(x_{0})$ \COMMENT{based on Algorithm.\ref{alg:draw_facepose_bbox}}
            \STATE $V_P = Visualization(x_{0})$
            \STATE $\mathbf{c}_{2} = \{I, V_M, V_P , E_a(a)\}$
            \STATE \textbf{Training}
            \REPEAT
                \STATE Sample $t \sim \text{Uniform}(\{1, \dots, T\})$
                \STATE $\mathcal{L}_{SMGA} \leftarrow \text{Loss}(D_{SMGA}(\mathbf{c}_{1}), x_{0})$ \hfill \COMMENT{based on Eq.\ref{eq:21}}
                \STATE $\theta_1 \leftarrow \theta_1 - \nabla_{\theta_1} (\mathcal{L}_{SMGA}, l_G)$        
                \STATE $\mathcal{L}_{G_{LV}} \leftarrow \text{Loss}(D_{LV}(\mathbf{c}_{2}), z_0)$ \hfill \COMMENT{based on Eq.\ref{eq:26}}
                \STATE $\theta_2 \leftarrow \theta_2 - \nabla_{\theta_2} (\mathcal{L}_{G_{LV}}, l_D)$
                \STATE $ = \theta_2 - \nabla_{\theta_2} \left\| \boldsymbol{\epsilon} - \boldsymbol{\epsilon}_{\theta_2} \left( \sqrt{\bar{\alpha}_t} z_0 + \sqrt{1 - \bar{\alpha}_t} \boldsymbol{\epsilon}, \mathbf{c}_{2}, t \right) \right\|^2$

            \UNTIL{converged}
            \STATE $id_{\min} \leftarrow \text{Argmin}([\text{rmse}_i])$ \hfill \COMMENT{the $id$ of the min values}
            \STATE $\theta_1^* \leftarrow [\theta_1^i](id_{\min}), \theta_2^* \leftarrow [\theta_2^i](id_{\min})$
        \end{algorithmic}
    \end{algorithm}
    \vspace{-2.5em}
\end{figure}

\section{EXPERIMENTS}

\subsection{Experimental Setup}
\textbf{Datasets.} 
Our study uses the PATS dataset\cite{ahuja2020style, ahuja-etal-2020-gestures, DBLP:journals/corr/abs-2007-12553}, which includes 84,000 clips from 25 speakers, averaging 10.7 seconds per clip and totaling 251 hours. We focused on a subset of four speakers (Jon, Kubinec, Oliver, and Seth), extracting 1,200 valid clips per speaker, for a total of 4,800 clips. Using a 0.5s step, we generated 3.2s clip, resulting in 58,844 training clips and 2,048 test clips.

\textbf{Evaluation Metrics.} The Fréchet Gesture Distance \textbf{(FGD)} \cite{FGD} is used to measure the distribution gap between real and generated gestures in the feature space, while Diversity \textbf{(Div.)} \cite{DIV} quantifies the variability of the generated gestures. Both metrics are calculated using an auto-encoder trained on PATS pose data.  For assessing video quality, we leverage Fréchet Video Distance \textbf{(FVD)}\cite{fvd}, kernel video distance \textbf{(KVD)}, and Fréchet Image Distance \textbf{(FID)} with the I3D classifier \cite{Joi3dcvpr17} pre-trained on Kinetics-400 \cite{Joi3dcvpr17}. Additionally, \textbf{PSNR}\cite{hore2010image}, \textbf{SSIM}\cite{wang2004image}, and \textbf{LPIPS} are employed to evaluate pixel-level similarity, ensuring that the generated video closely matches the ground truth. We also use  SyncNet\cite{Chung2016OutOT} to compute \textbf{Sync-C} and \textbf{Sync-D} to validate audio-lip synchronization accuracy.

\textbf{Implementation Details.}
We train our overall framework on four speakers jointly in two stages.\textbf{ (1) }For the SMGA-Network: We collect keypoints and expand them into pose sequence $x^i_0 \in \mathbb{R}^{402}$ for each frame. For audio feature extraction, we adopt the feature extraction strategy outlined in \cite{he2024co}, we set the batch size to 256, and use the Adam optimizer with a learning rate of \( 2 \times 10^{-4} \), a weight decay of 0.02, and 3,400 training epoch. The sampling step \( T \) is set to 50.  

\textbf{(2) }The multimodal video generation model is trained in two training processes. The first process involves 29,580 steps with a batch size of 2, while the second process includes 32,500 steps with a batch size of 1. Both processes have consistent parameter settings, including a video resolution of \( 512 \times 512 \). To improve the quality and robustness of video generation, all input modalities are randomly removed with a probability of 5\% during training. The learning rate for both stages is set to \( 1 \times 10^{-5} \), and optimization is done using the Adam optimizer.\cite{kingma2017adammethodstochasticoptimization}. All training experiments were conducted on a computing platform equipped with eight NVIDIA A6000 GPUs, whereas inference was carried out on a single NVIDIA A6000 GPU. The complete training procedure is summarized in Algorithm~\ref{alg2}. In addition, we report a comprehensive comparison of the computational overhead among our method and several representative baselines, including Animate-Anyone\cite{hu2024animate}, MimicMo\cite{zhang2024mimicmotion}, and Hallo\cite{xu2024hallo}. The results are summarized in Table~\ref{table:cpmputation}.

\begin{table*}[t]
\vspace{-0.5em}
\caption{\centering{Quantitative comparison of co-speech gesture video generation methods in terms of gesture, face, lip, and overall video quality metrics. This table compares the performance of our proposed method with baseline models (MM-Diffusion\cite{ruan2022mmdiffusion}, EDGE + Animate-anyone \cite{tseng2023edge, hu2024animate}, EDGE + Hallo\cite{tseng2023edge, xu2024hallo}, and DiffGest + MimicMo\cite{zhu2023taming, zhang2024mimicmotion} in audio-driven and pose-driven video generation tasks. The \textbf{bold text} indicates the optimal results, and the \underline{underlined text} indicates results that were obtained.}}
\vspace{-0.5em}
\centering
\setlength{\tabcolsep}{2pt} 
\renewcommand{\arraystretch}{1.15} 
\resizebox{1\textwidth}{!}{
\begin{tabular}{lcc|cccccc|cc|ccccc}
\toprule[1pt]
\multicolumn{1}{c}{Method} & \multicolumn{2}{c}{Gesture}  & \multicolumn{6}{c}{Face}&\multicolumn{2}{c}{Lip} & \multicolumn{5}{c}{Videos}\\
\cmidrule(lr){2-3} \cmidrule(lr){4-9} \cmidrule(lr){10-11}\cmidrule(lr){12-16} 
 & FGD($\downarrow$) & Div.($\uparrow$) & FID($\downarrow$) & FVD($\downarrow$)  & KVD($\downarrow$) & PSNR($\uparrow$) & SSIM($\uparrow$) & LPIPS($\downarrow$) & Syn-C($\downarrow$) & Syn-D($\uparrow$) & FID($\downarrow$) & FVD($\downarrow$) & PSNR($\uparrow$) & SSIM($\uparrow$) & LPIPS($\downarrow$) \\
\hline

MM-Diffusion\cite{ruan2022mmdiffusion}& 141.76 & \textbf{0.0998} & 39.28 & 693.31 & 63.150 & 27.94 & 0.1312 & 0.6112 & 13.75 & 0.82 & 86.18 & 1007.01 & 27.93 & 0.1667 & 0.7058  \\


EDGE + Animate-anyone\cite{tseng2023edge, hu2024animate} & \underline{6.72} & \underline{0.0855} & \underline{4.20} & \underline{176.92} & \underline{17.36} & \underline{29.76} & \underline{0.4188} & \underline{0.2670} & 12.27 & 2.01 & \textbf{6.91} & 410.92 & \underline{31.16} & 0.6152 & \underline{0.2374} \\

EDGE + Hallo\cite{tseng2023edge, xu2024hallo} & 7.57 & 0.0846 & 5.72 & 245.10 & 24.17 & 29.54 & 0.4158 & 0.2695 & \underline{12.09} & \underline{2.28} & 8.66 & 398.10 & 30.44 & 0.6056 & 0.2405 \\

DiffGest + MimicMo\cite{zhu2023taming, zhang2024mimicmotion} & 13.17 & 0.0854 & 5.78 & 239.53 & 18.37 & 29.68 & 0.4121 & 0.2837 & 12.48 & 1.21 & 9.40 & \underline{246.63} & 31.13 & \textbf{0.6408} & 0.2390 \\

Ours(Audio driven) & \textbf{6.02} & 0.0824 & \textbf{3.43} & \textbf{99.38} & \textbf{10.49} & \textbf{29.87} & \textbf{0.4413} & \textbf{0.2535} & \textbf{10.16} & \textbf{4.51} & \underline{7.90} & \textbf{230.89} & \textbf{31.29} & \underline{0.6378} & \textbf{0.2305} \\
\bottomrule[1pt]
\end{tabular}%
}\label{Tab:2}
\end{table*}

\begin{table*}[t]
\vspace{-0.7em}
\caption{\centering{Quantitative Comparison of Pose-Driven Gesture Video Generation Methods in Terms of Gesture, Face, Lip, and Overall Video Quality Metrics.} This table compares the performance of our proposed method with baseline models Animate-anyone \cite{hu2024animate} and MimicMo\cite{zhang2024mimicmotion} in pose video-driven video generation tasks.}
\vspace{-0.5em}
\centering
\setlength{\tabcolsep}{4pt} 
\renewcommand{\arraystretch}{1} 
\resizebox{1\textwidth}{!}{
\begin{tabular}{lcccccc|ccccc}
\toprule[1pt]
\multicolumn{1}{c}{Method} & \multicolumn{6}{c}{Face}& \multicolumn{5}{c}{Videos}\\
\cmidrule(lr){2-7} \cmidrule(lr){8-12} 
& FID($\downarrow$) & FVD($\downarrow$)  & KVD($\downarrow$) & PSNR($\uparrow$) & SSIM($\uparrow$) & LPIPS($\downarrow$) & FID($\downarrow$) & FVD($\downarrow$) & PSNR($\uparrow$) & SSIM($\uparrow$) & LPIPS($\downarrow$) \\
\hline

Animate-anyone\cite{hu2024animate}  & \underline{3.11} & 101.43 & 9.849 & \underline{31.50} & 0.8231 & 0.1018 & \underline{4.97} & 160.20 & \underline{32.51} & 0.8105 & 0.1252 \\

MimicMo\cite{zhang2024mimicmotion}  & 3.50 & \underline{94.13} & \textbf{6.136} & 31.39 & \textbf{0.8332} & \underline{0.0994} & 5.49 & \textbf{60.81} & 32.41 & \textbf{0.8181} & \underline{0.1228} \\

\textbf{Ours} & \textbf{2.96} & \textbf{84.24}  & \underline{8.531} & \textbf{31.59} & \underline{0.8315} & \textbf{0.0982} & \textbf{4.72} & \underline{135.00} & \textbf{32.62} & \underline{0.8168} & \textbf{0.1225} \\

\bottomrule[1pt]
\end{tabular}%
}\label{Tab:pose}
\vspace{-1.5em}
\end{table*}

\begin{table}[t]
  \centering
  \setlength{\tabcolsep}{13pt} 
  \renewcommand{\arraystretch}{0.6} 
  \caption{\centering{Sensitivity analyses concerning the loss weights of the first stage audio-to-pose model.}}
  \vspace{-0.5em}
  \label{Tab:loss}
  \resizebox{0.48\textwidth}{!}{
  \begin{tabular}{|l|cccc|}
  \toprule
  $\lambda_F : \lambda_B$ & \textbf{1:1} & \textbf{1:2 }&  \textbf{1:3} &  \textbf{1:4} \\
  \midrule
  FGD ($\downarrow$)& \textbf{6.705} & 7.296 & \underline{7.268} & 7.369 \\
  Syn-D($\uparrow$) & 4.308 & 4.268 & \textbf{4.409} & \underline{4.361}  \\
  \bottomrule
  \end{tabular}}
  \vspace{-1.5em}
\end{table}

\subsection{Quantitative Result}
To improve evaluation efficiency, we standardized the resolution of all videos to \( 256 \times 256 \) for assessment. A comprehensive evaluation of different metrics was conducted using a consistent test set to ensure fair and reliable comparisons.
Currently, open-source training code for co-speech gesture video generation methods is relatively limited, and the training code for some methods, such as \cite{hogue2024diffted}, \cite{chen2024echomimic}, \cite{zhang2024mimicmotion}, \cite{meng2024echomimicv2}, and \cite{corona2024vlogger}, is not publicly available, preventing direct comparisons on the same dataset. 
We built three two-stage co-speech gesture videos generation baselines by pairing gesture generators with video synthesizers: EDGE\cite{tseng2023edge} + Animate-anyone\cite{hu2024animate}, EDGE\cite{tseng2023edge} + Hallo\cite{xu2024hallo}, and DiffGest\cite{zhu2023taming} + MimicMo\cite{zhang2024mimicmotion}. EDGE\cite{tseng2023edge} adopts the S2G-MDD\cite{he2024co} strategy, using DWPose to extract keypoints from the PATS\cite{ahuja2020style, ahuja-etal-2020-gestures, DBLP:journals/corr/abs-2007-12553} corpus and pairing them with audio for training. EDGE + Hallo further incorporates Animate-anyone’s\cite{hu2024animate} pose-driven guider and replaces Hallo’s\cite{xu2024hallo} face-only static masks with region-specific upper-body masks produced by YOLOv5\cite{redmon2016you}.

\textbf{Co-speech Gesture Videos Generation.} As summarized in Table~\ref{Tab:2}, our MMGT establishes a well-balanced new state of the art. It reduces video-level FVD to 230.89, 6 \% lower than the previous best (DiffGest + MimicMo) and nearly half of EDGE + Hallo, while also delivering the highest PSNR and the lowest LPIPS. Its SSIM of 0.6378 is statistically tied for the lead. Lip-sync accuracy tops both measures (Syn-C, Syn-D). Facial quality is unmatched across all face metrics. Gesture realism is reflected in the best FGD. Although the Div metric may not perfectly reflect pose generation quality when significant jitter is present, it is less relevant in this context. This is further supported by the qualitative experiments in Fig.~\ref{fig:8}.

\textbf{Pose Video-driven Generation.} During inference, we replaced the audio-generated pose videos with full-body pose videos extracted from the PATS corpus using DWPose\cite{yang2023effective}. As shown in Table~\ref{Tab:pose}, our model (MMGT) delivers the strongest overall performance compared with Animate-anyone \cite{hu2024animate} and MimicMo \cite{zhang2024mimicmotion}. It achieves the top scores on nearly all facial metrics (FID, FVD, PSNR, LPIPS) with a competitive SSIM, while also leading the video-level FID, PSNR, and LPIPS, and noticeably reducing FVD relative to Animate-anyone. Qualitative comparisons in Fig.~\ref{fig:9} further demonstrate that MMGT maintains body structure and identity fidelity.

\subsection{Qualitative Results.}
\begin{table*}[t]
\caption{\centering{The quantitative results of the ablation study highlight the impact of the key modules on gesture generation, video quality and audiovisual synchronization metrics.}}
\vspace{-0.5em}
\centering
\setlength{\tabcolsep}{2pt} 
\renewcommand{\arraystretch}{1.15} 
\resizebox{1\textwidth}{!}{
\begin{tabular}{lcc|cccccc|cc|ccccc}
\toprule[1pt]
\multicolumn{1}{c}{\textbf{Method}} & \multicolumn{2}{c}{\textbf{Gesture}}  & \multicolumn{6}{c}{\textbf{Face}}&\multicolumn{2}{c}{\textbf{Lip}} & \multicolumn{5}{c}{\textbf{Videos}}\\
\cmidrule(lr){2-3} \cmidrule(lr){4-9} \cmidrule(lr){10-11}\cmidrule(lr){12-16}
\textbf{} & \textbf{FGD}($\downarrow$) & \textbf{Div.}($\uparrow$)& \textbf{FID}($\downarrow$)& \textbf{FVD}($\downarrow$) & \textbf{KVD}($\downarrow$) & \textbf{PSNR}($\uparrow$) & \textbf{SSIM}($\uparrow$) & \textbf{LPIPS}($\downarrow$) & \textbf{Syn-C}($\downarrow$) & \textbf{Syn-D}($\uparrow$) & \textbf{FID}($\downarrow$)& \textbf{FVD}($\downarrow$) & \textbf{PSNR}($\uparrow$) & \textbf{SSIM}($\uparrow$) & \textbf{LPIPS}($\downarrow$) \\
\hline
w/o SMT & 6.78 & 0.0813 & 17.99 & 262.20  & 25.01 & 29.42 &  0.3775 & 0.3250 & 11.83 & 2.25 & 17.91 & 409.53  & 30.42 & 0.5876 & 0.2735  \\

w/o FiLM & \underline{6.38} & \textbf{0.0860} & 4.51 & \underline{132.56} & \textbf{8.50} & \underline{29.85} & \underline{0.4400} & 0.2599 & 10.78 & 3.85 & \underline{9.11} & \underline{238.26} & \underline{31.25} & \underline{0.6321} & 0.2353 \\

w/o $\mathcal{L}_f$ & 7.09 & 0.0772 & 5.12 & 167.58 & \underline{16.40} & 29.45 & 0.3943 & 0.2877 & 11.16 & 3.18 & 11.49 & 411.42  & 30.54 & 0.6039 & 0.2544  \\

w/o Motion Mask & 6.51 & 0.0831 & \underline{4.24} & 205.08 & 29.70 & 29.57 & 0.4267 & 0.2618 & \underline{10.18} & \textbf{4.65} & 9.30 & 378.32 & 30.32 & 0.6108 & 0.2380 \\

w/o Audio & 6.72 & 0.0824 & 21.50 & 609.44 & 30.04 & 29.67 & 0.4315 & \underline{0.2553}  & 11.03 & 3.58 & 31.56 & 1250.71 & 30.80 & 0.6237 & \underline{0.2347}  \\

w Still Mask & 7.09 & \underline{0.0833} & 5.35 & 267.10  & 25.46 & 29.42 & 0.3775 & 0.3250 & 10.84 & 4.01 & 9.71 & 286.39  & 30.28 & 0.6050 & 0.2430 \\

Ours(Ours) & \textbf{6.02} & 0.0825 & \textbf{3.43} & \textbf{99.38} & 10.49 & \textbf{29.87} & \textbf{0.4413} & \textbf{0.2535} & \textbf{10.16} & \underline{4.51} & \textbf{7.90} & \textbf{230.89} & \textbf{31.29} & \textbf{0.6378} & \textbf{0.2305} \\

\bottomrule[1pt]
\end{tabular}%
}\label{Tab:3}
\vspace{-1.5em}
\end{table*}

\textbf{Qualitative Comparative Experiment.} 
Fig.~\ref{fig:8} gives a qualitative comparison of o-speech gestures video generation models on the PATS dataset. The baseline pipelines (MM-Difuusion, EDGE + Animate-anyone, EDGE + Hallo, and DiffGest + MimicMo) display delayed, unnatural hand motions and noticeable lip-sync errors, along with facial distortions and background artifacts. In contrast, the videos generated by our method are highly consistent with real data, while maintaining smooth gestures and time alignment, accurate lip-syncing, and no artefacts in the background.

We also provide a qualitative comparison of Pose-driven videos generation models on the PATS dataset, as shown in Fig.~\ref{fig:9}, where Animate-anyone\cite{hu2024animate} exhibits obvious hand-shape drift and palm artefacts (yellow boxes), and its mouth region intermittently blurs, breaking identity consistency. MimicMo\cite{zhang2024mimicmotion} alleviates some distortions but still shows wrist jitter and soft facial details. In contrast, the video clips generated by our model ensure that the wrist trajectory accurately follows the target posture. At the same time, facial expressions (including subtle movements of the eyes and lips) remain clear and stable over time.

\begin{figure}[t]
\centerline{\includegraphics[width=0.49\textwidth]{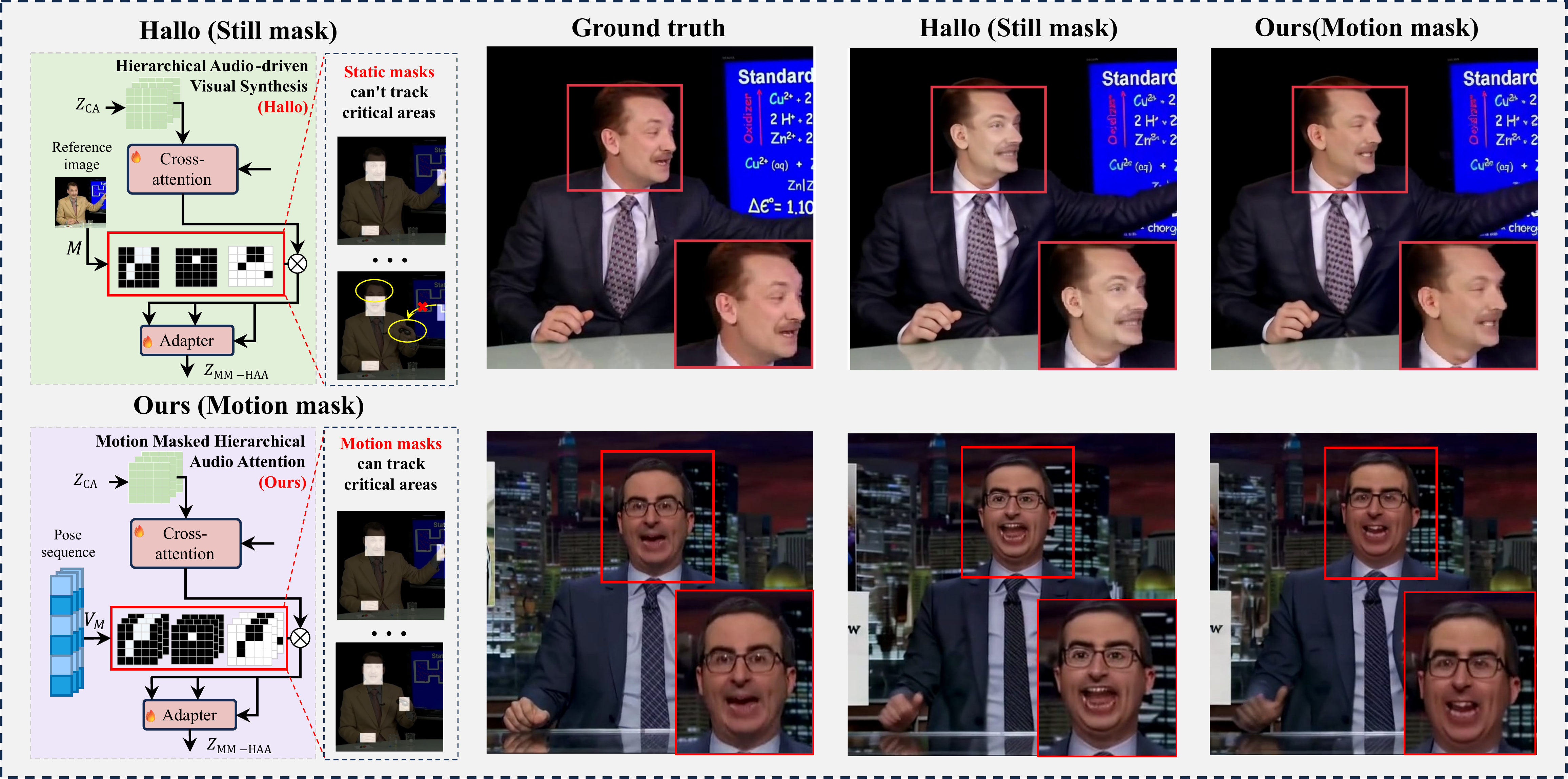}}
\caption{Qualitative Comparison of Video Generation with Still and Motion Masks.}
\vspace{-2.0em}
\label{fig:11}
\end{figure}

\textbf{Lips Synchronization with Audio.} For the qualitative comparison of lip synchronization, Fig.~\ref{fig:8} shows the lip synchronization results for the word 'negative' in the audio phrase. Our method (MMGT) generates lip shapes that closely match the Ground Truth, accurately capturing both the timing and subtle dynamics of the word. In contrast, EDGE + Animate-anyone \cite{tseng2023edge, hu2024animate}, DiffGest + MimicMo\cite{zhu2023taming, zhang2024mimicmotion}, and MM-Diffusion\cite{ruan2022mmdiffusion} display noticeable mismatches in lip movements, leading to poor synchronization with audio. EDGE + Hallo \cite{tseng2023edge, xu2024hallo} exhibits significant distortions and artifacts in the lips.
Additionally, we provide quantitative metrics for lip synchronization in Table~\ref{Tab:2} and Table~\ref{Tab:3}, which include Syn-C and Syn-D to evaluate synchronization accuracy and diversity, respectively. These results further confirm the robustness and superiority of our method over the baseline and ablation variants.
\begin{figure}[t]
{\includegraphics[width=0.5\textwidth]{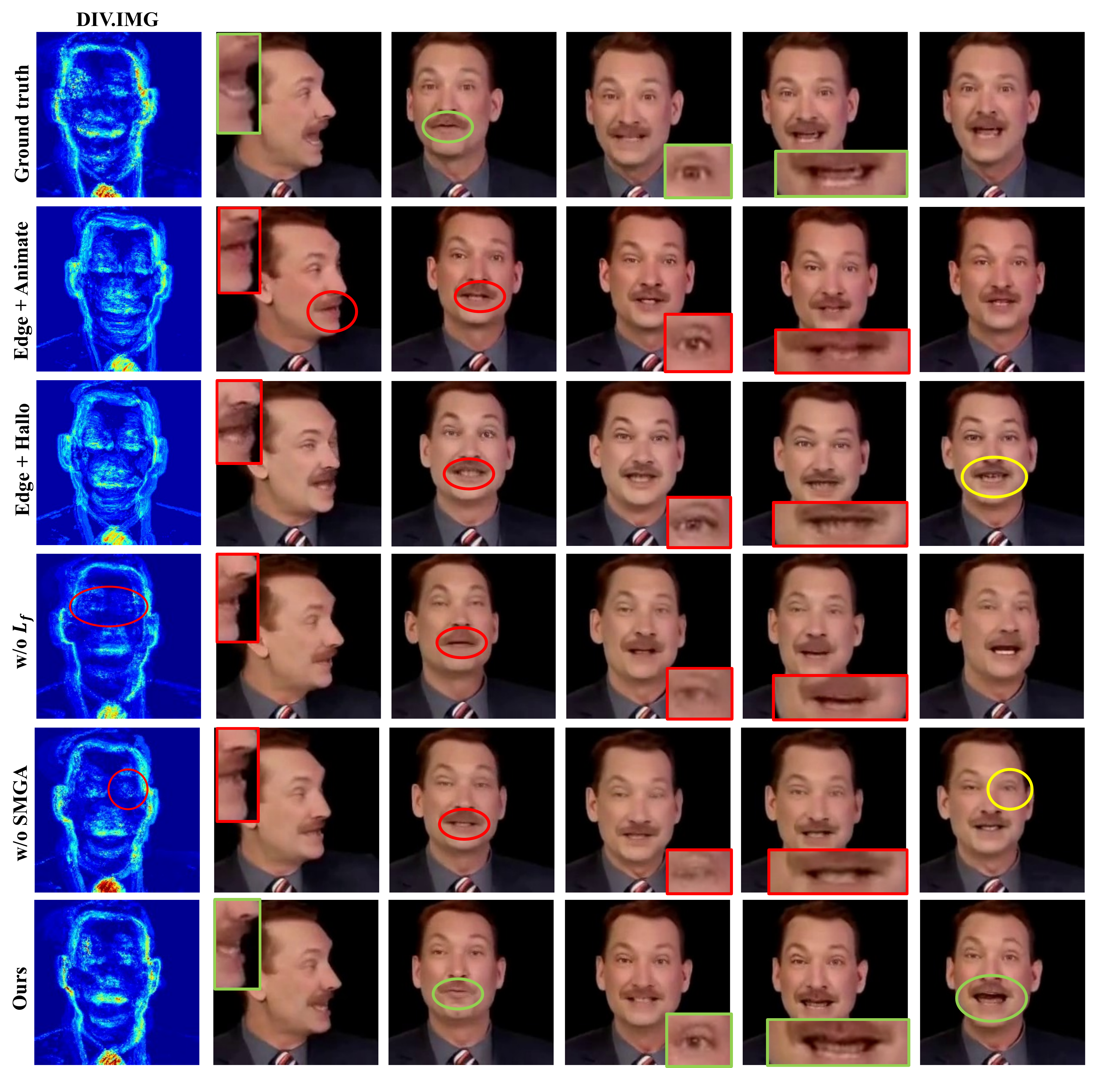}}
\caption{Presentation and qualitative comparison of overall head diversity maps based on the PATS dataset.}
\vspace{-1.5em}
\label{fig:12}
\end{figure}

\textbf{Effectiveness of Motion Masks in Video Generation.} 
Fig.~\ref{fig:11} shows a comparison of video generation results under the same input conditions using still masks and motion masks. Two key frames are selected to demonstrate the effectiveness of the proposed method. As shown, adding motion masks significantly improves the quality of generated videos, especially in the face and lip areas, where finer details and movement synchronization are better captured. By leveraging the dynamic information from motion masks, the model produces more realistic and coherent videos compared to using still masks. To give a thorough evaluation, we also confirm these findings with ablation experiments using quantitative metrics, as shown in Table~\ref{Tab:3}.


\textbf{Evaluation of Facial Motion Diversity and Synchronization.}
We performed a detailed comparison of facial expressions motion diversity and synchronization across individual test videos, evaluating Ground Truth, EDGE + Animate-anyone \cite{tseng2023edge, hu2024animate}, EDGE + Hallo \cite{tseng2023edge, xu2024hallo}, our method MMGT, and an ablation version of our method (MMGT w/o SMGA and MMGT w/o $\mathcal{L}_f$). As shown in Fig.~\ref{fig:12}, the motion diversity heatmap (DIV.IMG) on the left visualizes the magnitude of the motions, with blue indicating the smallest motions, green and yellow showing moderate motions, and red highlighting the most significant motions. From the comparison of the diversity maps, our method detects the most prominent and accurate lip movements that closely match the ground truth.


\textbf{Detailed Comparison of Hand Motion Across Methods.}
We highlight the performance of different methods in reproducing hand motion. As shown in columns 2, 5, and 8 of Fig.~\ref{fig:8}, MM-Diffusion \cite{ruan2022mmdiffusion} generates noticeable artifacts and blurred hand shapes, failing to capture the fine details and consistency of hand movements. Similarly, EDGE + Animate-anyone, EDGE + Hallo, and DiffGest + MimicMo struggle with accurate hand positioning and shape generation, resulting in significant differences from the Ground Truth. In contrast, our method demonstrates superior fidelity by accurately capturing realistic hand gestures with precise motion and clear structural details, closely aligning with the Ground Truth.

\begin{figure}[t]
\centerline{
\includegraphics[width=0.49\textwidth]{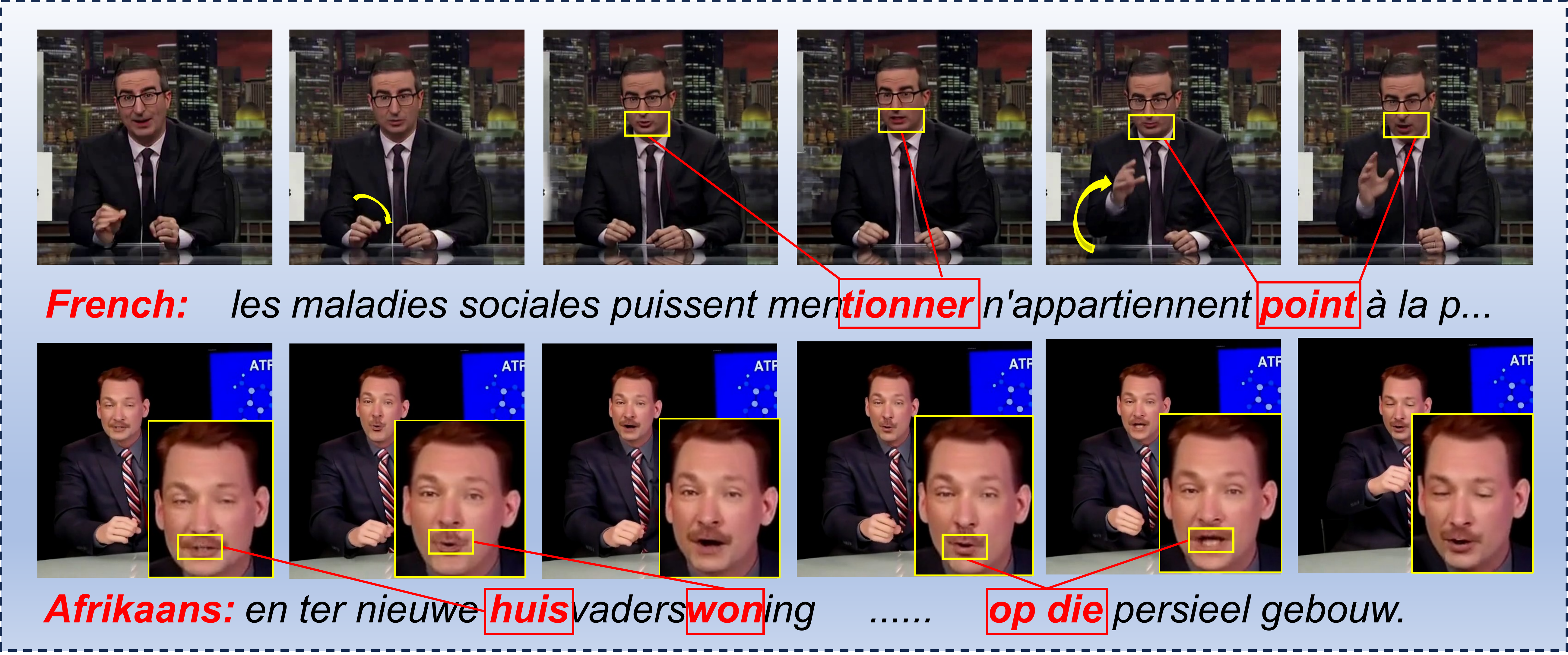}}
\caption{We test the generalisability of our model using languages such as French and Afrikaans, respectively.}
\label{fig:13}
\vspace{-1em}
\end{figure}

\textbf{Cross-lingual Generalization.}
If only French or Afrikaans speech is used to replace the audio track, and not paired video ground truths, MMGT still generates lip shapes and rhythmic hand gestures that match the prosody. Notably, changes in speaking rate are naturally reflected in the frequency of mouth closures and the duration of gesture pauses, suggesting that our model operates in a language-independent manner and is highly generalisable. Due to the lack of real videos in these languages, we currently present qualitative results, as shown in Fig.~\ref{fig:13}, and plan to evaluate them manually in future work.

\begin{figure}[t]
\centerline{\includegraphics[width=0.49\textwidth]{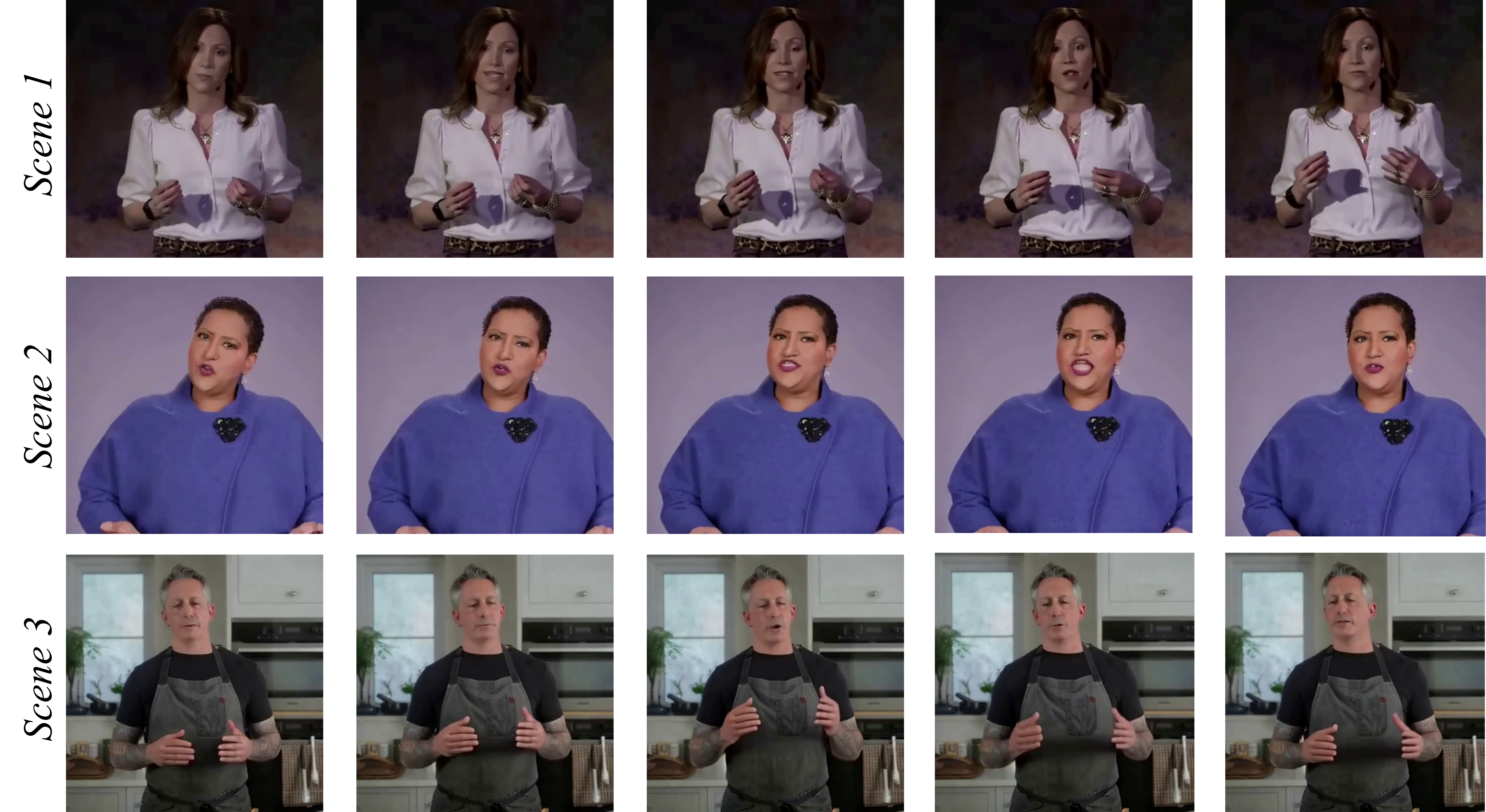}}
\caption{MMGT(Ours) supports many different scenarios and unseen speakers.}
\vspace{-1.5em}
\label{fig:14}
\end{figure}

\textbf{Generalisation to Unseen Speakers and Multiple Scenes.} We trained MMGT using the TED-72h video dataset and evaluated it on unseen speakers. As shown in Fig.~\ref{fig:14}, the three rows correspond to indoor natural light, stage lighting, and a kitchen scene with deep depth of field, respectively. Under these heterogeneous backgrounds and camera positions, MMGT is able to preserve clothing textures, skin tones, and gestures; images remain stable, and facial details are consistent, indicating that the technique can be directly applied to real-world scenes and unseen speakers.

\subsection{Ablation Study.}

Table~\ref{Tab:3} reports a comprehensive ablation in which we disable, one at a time, the major architectural and training ingredients of MMGT. Metrics cover gesture realism (FGD ↓, Div. ↑), facial quality, lip–audio alignment, and full-video fidelity.  

\textbf{Spatial Mask-guided Two-branch structure (SMT).} Removing SMT (“w/o SMT”) causes the sharpest overall degradation. FGD rises from 6.02→6.78 and facial FVD more than doubles (99.38→262.20), while Syn-C/Syn-D indicate poorer lip timing (11.83 ↑ / 2.25 ↓). These drops confirm SMT’s role in coordinating coarse body motion with fine facial cues. Gesture and facial quality also decreased, as evidenced by higher FGD and FVD scores, highlighting the importance of SMGA in maintaining robust lip synchronization (Fig.~\ref{fig:11} and Fig.~\ref{fig:12}).

\textbf{Feature-wise Linear Modulation (FiLM).} Excluding FiLM (“w/o FiLM”) moderately harms most face–video measures (e.g., facial FVD 99.38→132.56, video FVD 230.89→238.26), even though gesture diversity Div. improves slightly. The result suggests FiLM chiefly refines appearance rather than motion amplitude.

\textbf{Facial Motion Loss $\mathcal{L}_f$.} Similarly, substituting the motion-specific facial loss \(\mathcal{L}_f\) with a generalized loss leads to a pronounced decrease in lip synchronization accuracy. The Syn-C metric rises from 10.16 to 11.11, and the Syn-D metric increases from 4.51 to 4.65, signifying diminished articulation precision and temporal coherence between the lips and audio.  As shown in Fig.~\ref{fig:12}, this degradation visibly manifests as weaker lip movements aligned with the spoken content.

\textbf{Motion Mask in MM-HAA.} Replacing the dynamic mask with none (“w/o Motion Mask”) boosts Syn-D to the best value (4.65), but facial and video FVDs climb sharply (205.08 and 378.32), showing that the mask primarily stabilizes global coherence and texture. 

\textbf{Audio Guidance in Stage II.} Removing the audio input entirely and relying solely on keypoint-driven generation leads to substantial performance drops in both facial and overall metrics (e.g., FVD and FID), emphasizing the essential role of audio in enhancing visual quality and temporal consistency. Furthermore, the notable decrease in lip synchronization highlights the necessity of audio-based guidance for achieving realistic, accurate lip movements.

\textbf{Replace Motion Mask by Still Mask.} When a still mask replaces the motion mask in the MM-HAA module, face-specific metrics such as SSIM and PSNR decrease (from 0.3945 to 0.3775 and from 29.47 to 29.42, respectively), while the facial FVD score increases from 99.38 to 267.10, indicating a marked loss of detail in the face region. Gesture coherence also degrades (FGD rises from 7.04 to 7.089), and the overall video FVD increases from 230.89 to 286.389, pointing to diminished temporal consistency and video quality (Fig.~\ref{fig:10}).

With all components active, MMGT attains the statistically tied best scores on 13 of 16 indicators (e.g., facial FID 3.43, video PSNR 31.29 dB, Syn-C 10.16), validating that each module contributes to natural, temporally consistent, and visually faithful co-speech gesture video synthesis.

\subsection{Ethical Considerations and Mitigation Strategies}
This study highlights two major societal risks in audio-driven portrait animation technology: potential misuse for creating deceptive deepfakes and privacy breaches through unauthorized use of personal biometric data. Addressing these issues requires establishing ethical standards, ensuring transparent implementation, obtaining clear user consent, and enforcing strong data governance. These steps aim to align technological progress with social responsibility while encouraging socially beneficial uses.

\begin{figure}[t]
\centerline{\includegraphics[width=0.49\textwidth]{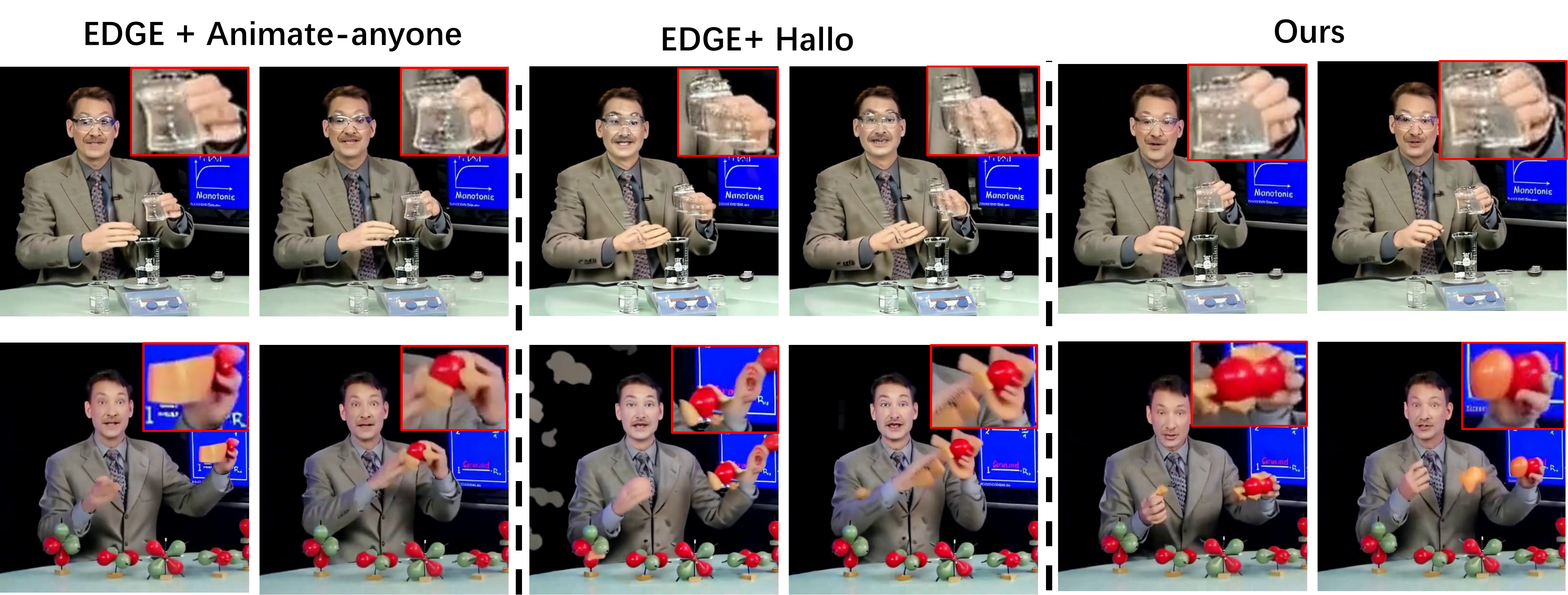}}
\caption{Qualitative Comparison of Collaborative Speech Gesture Video Generation Methods in the Face of Complex Interactions.}
\vspace{-1.5em}
\label{fig:15}
\end{figure}

\section{Discussion and limitations}

Although MMGT significantly improves overall motion realism and texture fidelity, it still has shortcomings when handling fine physical interactions (such as hand-to-face contact or hand-to-object manipulation). As shown in Fig.~\ref{fig:15}, we found that this phenomenon is common in existing methods\cite{hu2024animate, xu2024hallo}. We believe that this defect may mainly stem from the lack of dense annotated training data for such events and the lack of explicit contact constraints in the current architecture.

\section{Conclusions.}
We propose a Motion Mask-Guided Two-Stage Network (MMGT) for Co-Speech Gesture Video Generation, which combines motion masks with features extracted from audio. The Spatial Mask-Guided Audio2Pose Generation (SMGA) network captures large-scale facial and hand gesture movements, while the Motion Masked Hierarchical Audio Attention (MM-HAA) mechanism refines fine details within a stabilized diffusion framework. Experiments demonstrate that MMGT produces natural, high-quality videos with improved motion realism and texture accuracy.

\bibliographystyle{IEEEtran} 
\bibliography{references} 

\vspace{-2em}
\begin{IEEEbiography}
[{\includegraphics[width=1in,height=1.25in, clip, keepaspectratio]{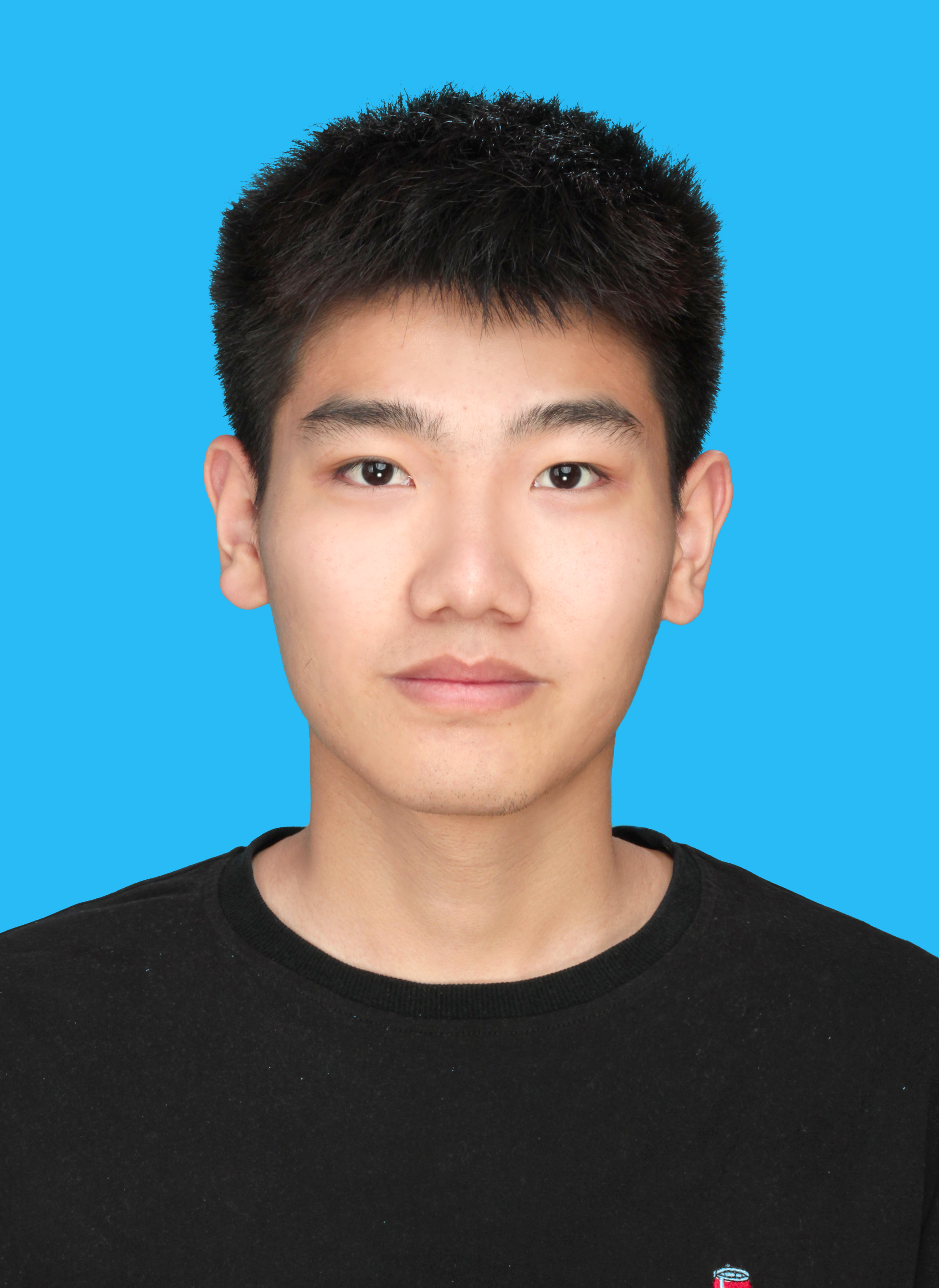}}]{Siyuan Wang}(Student Member, IEEE) received his B.E. degree from Wuhan University of Technology in 2023. 

Currently, he is an M.E. student at Shenyang Institute of Automation, Chinese Academy of Sciences. His current research interests include video generation, human pose estimation, and deep learning.
\end{IEEEbiography}
\vspace{-3em}


\begin{IEEEbiography}[{\includegraphics[width=1in,height=1.25in,clip,keepaspectratio]{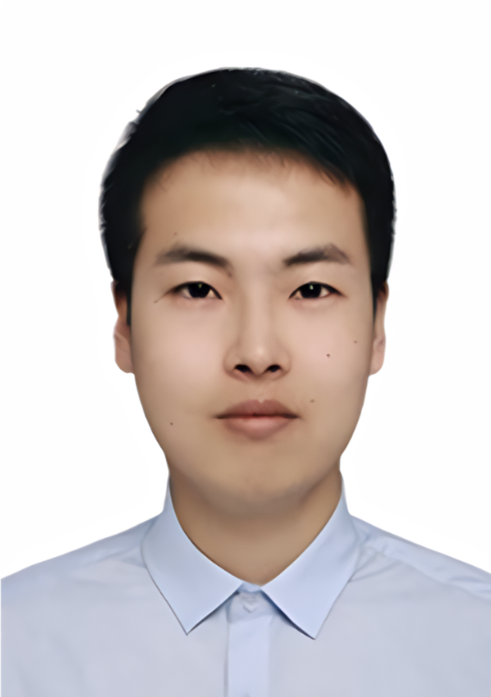}}]{Jiawei Liu}
(Member, IEEE) received his B.E. degree from Northeast Agricultural University, Harbin, China, in 2018, and his Ph.D. degree in Pattern Recognition and Intelligent Systems from the Shenyang Institute of Automation, Chinese Academy of Sciences, Shenyang, China, in 2024. 

He is currently a research assistant professor at Shenyang Institute of Automation, Chinese Academy of Sciences. His research interests include deep learning, illumination processing, image restoration, shadow removal, and diffusion models.
\end{IEEEbiography}
\vspace{-3em}

\begin{IEEEbiography}[{\includegraphics[width=1in,height=1.25in,clip,keepaspectratio]{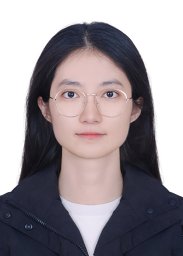}}]{Yeying Jin}
(Member, IEEE) received her B.E. degree from the University of Electronic Science and Technology of China (UESTC), and received her M.Sc. and Ph.D. degrees from the National University of Singapore (NUS). 

She is currently a Senior Researcher at Tencent and previously completed a research internship at Adobe, where she was mentored by Prof. Connelly Barnes. Her research interests include computer vision and deep learning, with a focus on image and video generation and enhancement.
\end{IEEEbiography}
\vspace{-3em}

\begin{IEEEbiography}[{\includegraphics[width=1in,height=1.25in, clip,keepaspectratio]{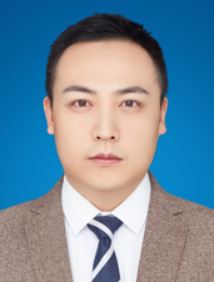}}]{Wei Wang} received his B.E. degree and M.E. degree from Xidian University, Xi’an, China. In 2014, he received a doctorate degree from the Shenyang Institute of Automation, Chinese Academy of Sciences, China. 

He is currently a professor at the Shenyang Institute of Automation, Chinese Academy of Sciences, and deputy director of the Intelligent Inspection and Equipment Research Laboratory.
\end{IEEEbiography}
\vspace{-3em}
\begin{IEEEbiography}[{\includegraphics[width=1in,height=1.25in, clip,keepaspectratio]{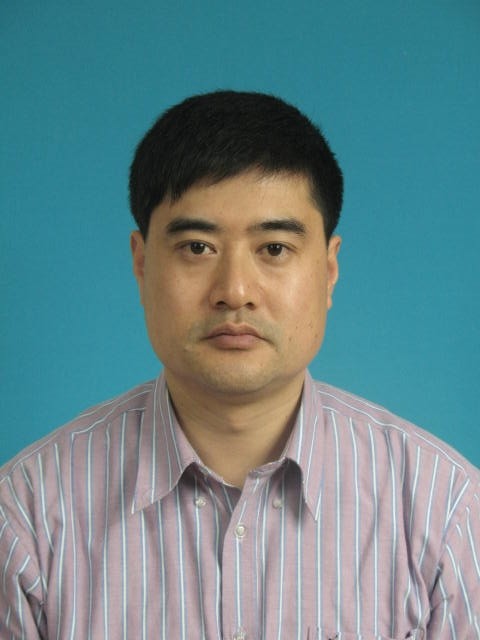}}]{Jinsong Du} received his B.E. degree from Shenyang University of Technology, Shenyang, China, in 1993, and his Ph.D. degree in Chinese Academy of Sciences (CAS) Graduate School, Chinese Academy of Sciences, Shenyang, China, in 2010. 

He is currently a professor at the Shenyang Institute of Automation, Chinese Academy of Sciences, where he serves as director of the Intelligent Inspection and Equipment Research Laboratory.
\end{IEEEbiography}
\vspace{-3em}
\begin{IEEEbiography}[{\includegraphics[width=1in,height=1.25in,clip,keepaspectratio]{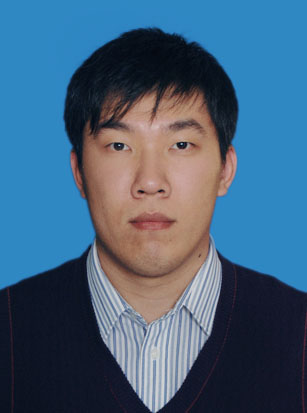}}]{Zhi Han}(Member, IEEE) 
received the B.Sc., M.Sc., and Ph.D. degrees in applied mathematics from Xi’an Jiaotong University (XJTU), Xi’an, China, in 2005, 2007, and 2012, respectively, and the joint Ph.D. degree in statistics from the University of California, Los Angeles (UCLA), Los Angeles, California, in 2011. 

He is currently a professor at the State Key Laboratory of Robotics, Shenyang Institute of Automation, Chinese Academy of Sciences (SIA, CAS). His research interests include image/video modeling, low-rank matrix recovery, and deep neural networks.
\end{IEEEbiography}
\end{document}